\documentclass[manuscript]{acmart}
\usepackage{graphicx}
\usepackage{url}
\usepackage{appendix}

\AtBeginDocument{%
  }

\begin{document}

\title{Graph Learning in the Era of LLMs: A Survey from the Perspective of Data, Models, and Tasks}



\author{Xunkai Li}
\email{cs.xunkai.li@gmail.com}
\affiliation{%
  \institution{Beijing Institute of Technology}
  \state{Beijing}
  \country{China}
}
\author{Zhengyu Wu}
\email{jeremywzy96@outlook.com}
\affiliation{%
  \institution{Beijing Institute of Technology}
  \state{Beijing}
  \country{China}
}
\author{Jiayi Wu}
\email{158937299@qq.com}
\affiliation{%
  \institution{Beijing Institute of Technology}
  \state{Beijing}
  \country{China}
}
\author{Hanwen Cui}
\email{2228618603@qq.com}
\affiliation{%
  \institution{Beijing Institute of Technology}
  \state{Beijing}
  \country{China}
}
\author{Jishuo Jia}
\email{jishuojia447@gmail.com}
\affiliation{%
  \institution{Shandong University}
  \state{Shandong}
  \country{China}
}
\author{Ronghua Li}
\email{lironghuabit@126.com}
\affiliation{%
  \institution{Beijing Institute of Technology}
  \state{Beijing}
  \country{China}
}

\author{Guoren Wang}
\email{wanggrbit@gmail.com}
\affiliation{%
  \institution{Beijing Institute of Technology}
  \state{Beijing}
  \country{China}
}

\begin{abstract}
    With the increasing prevalence of cross-domain Text-Attributed Graph (TAG) {Data} (e.g., citation networks, recommendation systems, social networks, and ai4science), the integration of Graph Neural Networks (GNNs) and Large Language Models (LLMs) into a unified {Model} architecture (e.g., LLM as enhancer, LLM as collaborators, LLM as predictor) has emerged as a promising technological paradigm. 
    The core of this new graph learning paradigm lies in the synergistic combination of GNNs' ability to capture complex structural relationships and LLMs' proficiency in understanding informative contexts from the rich textual descriptions of graphs.
    Therefore, we can leverage graph description texts with rich semantic context to fundamentally enhance {Data} quality, thereby improving the representational capacity of model-centric approaches in line with data-centric machine learning principles. 
    By leveraging the strengths of these distinct neural network architectures, this integrated approach addresses a wide range of TAG-based {Task} (e.g., graph learning, graph reasoning, and graph question answering), particularly in complex industrial scenarios (e.g., supervised, few-shot, and zero-shot settings).
    In other words, we can treat text as a medium to enable cross-domain generalization of graph learning {Model}, allowing a single graph model to effectively handle the diversity of downstream graph-based {Task} across different data domains.
    This survey offers a comprehensive review of graph learning in the era of LLMs, introducing a systematic summary built around the three pillars of machine learning: {Data}, {Model}, and {Task}.
    It examines the unique potential of LLM-GNN integration in handling heterogeneous {Data} sources, utilizing advanced neural {Model} architectures, and tackling various downstream {Task} in practical applications. 
    Categorizing existing studies within this comprehensive survey, we highlight key trends, reveal current challenges, and propose directions for future research. 
    This work serves as a foundational reference for researchers and practitioners looking to advance graph learning methodologies in the rapidly evolving landscape of LLM.
    We consistently maintain the related open-source materials at \url{https://github.com/xkLi-Allen/Awesome-GNN-in-LLMs-Papers}.
\end{abstract}


\maketitle

\section{Introduction}
   Graphs are an effective means of capturing the structural relationships between entities by representing nodes as individual entities and edges as their interrelationships.
   This makes graphs an ideal representation for a wide range of fields, including social networks, recommendation systems, and molecular structures, all of which involve complex interdependencies that are challenging to model using traditional data structures.
   Despite their power, graph data is inherently complex and heterogeneous, posing significant challenges for conventional methods. 
   Specifically, traditional approaches often fail to fully capture the intricate structural patterns within graph data, limiting their ability to model complex relationships and interactions effectively. 
   These limitations are exacerbated when dealing with large-scale graphs, where relationships between nodes are not only context-dependent but also exhibit considerable variability across different domains.
    
    In response to these challenges, Graph Neural Networks (GNNs) have emerged as a transformative tool. 
    Specifically, GNNs overcome the limitations of traditional methods by employing recursive message-passing mechanisms, which enable information to propagate across the graph structure. 
    Through this iterative process, GNNs can capture long-range dependencies and complex structural relationships, even in large and intricate graphs, allowing them to generate rich embeddings for nodes, edges, and entire graphs. 
    These learned embeddings empower GNNs to effectively support a wide range of graph-based learning tasks, such as node classification, link prediction, and graph classification. 
    The ability of GNNs to process complex graph data in an end-to-end, scalable manner has solidified their position as a core technique in graph machine learning research.

    Recently, the rise of Large Language Models (LLMs) has significantly advanced the field of artificial intelligence, demonstrating exceptional capabilities in natural language understanding, generation, and the modeling of complex patterns from vast datasets. 
    Simultaneously, the concept of data-centric machine learning has gained widespread attention, prompting researchers in the graph machine learning community to focus on enhancing graph representations from a data-driven perspective. 
    This shift has led to the proliferation of Text-Attributed Graphs (TAGs) across various domains, as the rich semantic information embedded in graph descriptions can further refine graph representations and improve data quality in many practical applications.
    
    Building on this progress, the integration of LLMs with graph learning has emerged as a transformative approach, offering new opportunities to push the boundaries of graph learning.
    As a general-purpose machine learning framework, LLMs provide unparalleled capabilities that can significantly enhance graph learning in ways previously unattainable. 
    The motivation for combining LLMs with graph learning can be understood from three key perspectives: \textbf{Data}, \textbf{Model}, and \textbf{Task}. 
    This integration not only improves graph data representation but also enables the development of more flexible models and strengthens reasoning capabilities for various graph-based tasks.
    
    From a data perspective, the increasing availability of high-quality, multi-domain TAGs has created more enriched, and diversified datasets. 
    LLMs, with their powerful natural language understanding and generation capabilities, can effectively capture semantic features from the text, which GNNs can then leverage to enhance the representational power of graph models. 
    These features, which provide a deep understanding of node and edge semantics, are crucial for improving graph models' ability to learn complex, context-sensitive relationships.
    Leveraging LLMs to extract high-quality semantic features marks the advancement in data-centric machine learning, and such integration pushes the boundaries of model accuracy and generalization further.
    
    From a model perspective, the collaboration between LLMs and GNNs offers a promising approach to overcoming the limitations of traditional graph learning techniques. 
    Researchers are increasingly exploring joint training and inference methods that combine the strengths of both models. 
    For instance, by using node attributes, such as textual information, LLMs can enhance GNN encodings, enabling better cross-domain generalization and reducing the need for domain-specific retraining. 
    This synergy not only improves model generalization but also addresses the common issue of poor transferability in applications requiring more flexible, broad-reaching models, thereby enhancing the overall versatility of graph learning.
    
    From a task perspective, the rise of complex industrial applications requires models that can effectively reason across diverse and large-scale data sources. 
    LLMs, with their exceptional ability to understand context, provide powerful tools for few-shot and zero-shot inference, which are critical in dynamic, real-world settings. 
    By combining the reasoning power of LLMs with the structural learning capabilities of GNNs, this integrated approach enables models to tackle a wider range of tasks in graph learning, from node classification to link prediction, with enhanced inference abilities that are adaptive to new, unseen tasks.

    In summary, the integration of LLMs with graph learning effectively addresses key challenges in graph machine learning, enhancing data representation, improving generalization, and enabling advanced reasoning across a range of tasks and application scenarios. 
    The synergy between LLMs and graph learning has the potential to unlock unprecedented performance and applicability, enabling more robust and scalable solutions for complex graph-based tasks. 
    This survey reviews the intersection of GNNs and LLMs, highlighting recent advancements, applications, and challenges, with a focus on their integration. 
    We propose a systematic framework to categorize current research, highlighting key trends, limitations, and future research directions in this rapidly evolving field. To help readers quickly understand the methods included and the various types of graph datasets, we provide dataset details and method descriptions.

\section{Why GNN needs LLM: The Role of Data}
GNNs have made significant advancements in processing structured graph data. However, when handling with graphs enriched with textual attributes, i.e., text-attributed graphs, traditional GNNs struggle to manage excessive textual information. 
In applications including recommendation systems, citation networks, and social network analysis, correspondent text-attributed graphs contain a larger amount of textual information compared to conventional graph datasets.
Textual information is vital for improving the performance of graph learning tasks, but existing GNN models are generally unable to process text data directly. Consequently, a key challenge is how to effectively extract features from text and integrate them into graph learning processes to enhance model performance.

Traditional text feature extraction methods, such as TF-IDF and Word2Vec, while useful in some applications, have significant limitations when dealing with long texts and complex contextual relationships. These methods primarily rely on fixed text vector representations that do not capture the deeper hierarchical and context-dependent information inherent in the node-edge structure of the graph. As a result, they struggle to handle sophisticated linguistic or graph-based tasks. 
In contrast, LLMs, which are trained on vast amounts of text, excel at capturing long-range dependencies and contextual semantic relationships. LLMs can generate more nuanced and high-quality text embeddings, overcoming the limitations of traditional methods. By leveraging tasks such as text generation, question answering, and sentiment analysis, LLMs provide richer and more precise feature representations for graphs. This enables more effective integration of text and graph data, leading to advancements in graph learning.
As a result, the integration of GNNs and LLMs has led to new breakthroughs in graph learning tasks.

We categorize existing research into five distinct groups based on data characteristics and the generalizability of methods across different graph-learning tasks. This approach is both intuitive and effective, as it addresses two key dimensions of graph-based research from the data perspective: graph-learning tasks and application domains.

This framework provides a clear assessment of methods’ versatility across various contexts. Grouping studies by tasks such as node classification or graph generation reveals how methods handle varying complexities, distinguishing between task-specific and more task-agnostic approaches. Categorizing by domains, such as social networks or biological databases, shows how methods are tailored to field-specific needs or generalized across applications.

    From the data perspective, LLMs enhance GNNs by providing high-quality feature representations. 
    This integration boosts GNN performance across diverse domains, enabling more effective handling of complex, context-rich information.  
    On the other hand, LLMs can benefit from GNNs' ability to capture relational and structural patterns.

\begin{itemize}
 
  \item \textbf{Single-task \& Single-domain.} This category focuses on studies that address specific tasks and domains within a unified dataset domain. These studies typically examine the direct application of GNNs and LLMs to simpler, well-defined graph structures, where the relationships between nodes and edges are clearly established within a consistent semantic space. Currently, the main works in this category include: LPNL\cite{bi2024lpnlscalablelinkprediction}, DGTL\cite{qin2024disentangledrepresentationlearninglarge}, AUGGLM\cite{xu2024languagemodelsgraphlearners}, GQT\cite{wang2024learninggraphquantizedtokenizers}, Link2Doc\cite{Link2Doc}, RoSE\cite{seo2024unleashingpotentialtextattributedgraphs}, TAGA\cite{zhang2024tagatextattributedgraphselfsupervised}, TAGExplainer\cite{pan2024tagexplainernarratinggraphexplanations}, CIKM-KD\cite{pan2024distillinglargelanguagemodels}, CurGL\cite{anonymous2024curriculum}, DP-GPL\cite{anonymous2024dpgpl}, GPPT\cite{gppt}, GRAD\cite{mavromatis2023traingnnteachergraphaware}, GraphBridge\cite{anonymous2024graphbridge}, GIANT\cite{chien2022nodefeatureextractionselfsupervised}, GLEM\cite{zhao2023learninglargescaletextattributedgraphs}, LLM-GNN\cite{chen2024labelfreenodeclassificationgraphs}, TAPE\cite{he2024harnessingexplanationsllmtolminterpreter}, ENGINE\cite{zhu2024efficienttuninginferencelarge}, GAugLLM\cite{fang2024gaugllmimprovinggraphcontrastive}, HiGPT\cite{tang2024higptheterogeneousgraphlanguage}, LMGJoint\cite{LMGJoint}, GPF\cite{fang2023universal}, SheaFormer\cite{SheaFormer}, G2P2\cite{G2P2}, SKETCH\cite{SKETCH}, GraphAdapter\cite{huang2024gnngoodadapterllms}, GraphPrompter\cite{graphprompter}, ZeroG\cite{li2024zeroginvestigatingcrossdatasetzeroshot}.

  \item \textbf{Single-task \& Multi-domain}. This category focuses on studies that apply GNNs and LLMs to single tasks across multiple domains. These studies highlight how these models can generalize to various types of graphs, leveraging domain-specific textual information from diverse sources. By demonstrating the ability to process heterogeneous graph structures and domain-specific textual features, these methods provide a powerful framework for tackling graph-learning tasks in a wide range of real-world applications. The representative research papers include the following: GraphAny\cite{zhao2024graphanyfoundationmodelnode}, GraphFM\cite{anonymous2024graphfm}, GraphProp\cite{liu2023graphpromptunifyingpretrainingdownstream}.

  \item \textbf{Multi-task \& Single-domain}. This category simultaneously handles multiple tasks within a single domain. These studies illustrate the potential of multi-task learning frameworks, which leverage shared representations for improving model efficiency and performance across various graph tasks. The research work reflecting the current state of the field mainly includes: PATTON\cite{jin2023pattonlanguagemodelpretraining}, ConGraT\cite{brannon2024congratselfsupervisedcontrastivepretraining}, SimTeG\cite{duan2023simtegfrustratinglysimpleapproach}, FineMolTex\cite{li2024finemoltexfinegrainedmoleculargraphtext}, HIGHT\cite{chen2024highthierarchicalgraphtokenization}, PGT\cite{pgt}, InstructGLM\cite{Instruct-GLM}, Grenade\cite{li2023grenadegraphcentriclanguagemodel}, THLM\cite{THLM}, GAGA\cite{GAGA}, LLaGA\cite{chen2024llagalargelanguagegraph}, Mol-MVMoE\cite{MolMvMOE}, GraphGPT\cite{tang2024graphgptgraphinstructiontuning}, GaLM\cite{xie2023graphawarelanguagemodelpretraining}, ZeroG\cite{li2024zeroginvestigatingcrossdatasetzeroshot}, PromptGFM\cite{PromptGFM}, TouchUp-G\cite{zhu2023touchupgimprovingfeaturerepresentation}, GraphTranslator\cite{zhang2024graphtranslatoraligninggraphmodel}, GraphText\cite{zhao2023graphtextgraphreasoningtext}, All in One\cite{AllInOne}, OMOG\cite{OMOG}.

  \item \textbf{Multi-task \& Multi-domain} This category extends the previous category to multiple domains.
  The goal of these studies are creating more generalizable graph learning models—Graph Foundation Model—that can handle diverse types of graph data and tasks across various application domains. These papers are the following: InstructGraph\cite{wang2024instructgraphboostinglargelanguage}, GPT4Graph\cite{guo2023gpt4graphlargelanguagemodels}, AnyGraph\cite{xia2024anygraphgraphfoundationmodel}, GOFA\cite{GOFA}, SCORE\cite{SCORE}, OpenGraph\cite{xia2024opengraphopengraphfoundation}, GFSE\cite{anonymous2024gfse}, GIT\cite{GIT}, GL-Fusion\cite{anonymous2024glfusion}, GraphBridge\cite{anonymous2024graphbridge}, OFA\cite{OFA}, PRODIGY\cite{huang2023prodigy}, WalkLM\cite{WalkLM}, OMOG\cite{OMOG}, MuseGraph\cite{tan2024musegraphgraphorientedinstructiontuning}, RAGraph\cite{jiang2024ragraphgeneralretrievalaugmentedgraph}, GraphAlign\cite{hou2024graphalignpretraininggraphneural}, EdgePromp\cite{anonymous2024edge}, GraphPrompt\cite{liu2023graphpromptunifyingpretrainingdownstream}, MultiGPrompt\cite{yu2024multigpromptmultitaskpretrainingprompting}.
  
  \item \textbf{Graph Reasoning} This category delves into the inference process within graph learning, focusing on how GNNs and LLMs can be used to infer new relationships or predict missing information in a graph. These studies represent advanced research that explores the mechanisms through which models make predictions and derive insights from complex graph structures. The main works in this category include: GraphLLM\cite{chai2023graphllmboostinggraphreasoning}, GraphAgent-Reasoner\cite{hu2024scalableaccurategraphreasoning}, GraphToken\cite{perozzi2024letgraphtalkingencoding}, Talk like a Graph\cite{fatemi2023talklikegraphencoding}, NLGraph\cite{NLGraph}.

\end{itemize}

We provide representative research papers that illustrate the current state of the field for each of these categories. These works serve as benchmarks for understanding how GNNs and LLMs are being applied and refined to address the challenges of learning from text-attributed graphs. Based on these papers, it is evident that research focusing on Single-task or Single-domain approaches faces significant limitations in terms of generalization capability and application scope. In contrast, Multi-task \& Multi-domain methods not only achieve broader applicability across diverse application scenarios but also significantly enhance overall model performance through shared underlying representations and collaborative learning among tasks. Therefore, future research should concentrate on Multi-task \& Multi-domain approaches, leveraging Graph Reasoning techniques to deeply understand and infer complex graph structures and node relationships, thereby progressively realizing Graph Foundation Model.

\section{Overview of Applied Datasets by Included Methods}
\label{Applied Datasets}

This section introduces the datasets used in the evaluated methods, organizing them into five domains for improved clarity and structure. From a data-centric perspective, categorizing datasets by domain is crucial for emphasizing their unique characteristics, such as data structure, scale, and task relevance. It also helps readers understand which specific domains the methods are intended to apply to.  Table~\ref{tab: Citation Network and Wikipage derived datasets} includes datasets that share a common characteristic of modeling interconnections between pieces of knowledge. The Citation Network captures relationships between scientific publications, typically sourced from DBLP and Arxiv, while Wikipedia and Web Page datasets represent pages of information regarding real-world entities as nodes, with edges capturing interactions between them.

Table~\ref{tab: E-commerce Datasets} presents datasets derived from e-commerce platforms, which typically include user-item interactions and are commonly used in recommendation system tasks. This category is well-established in the literature and represents a distinct domain with clear applications in personalized recommendations and user behavior analysis. Table~\ref{tab: Social Network Datasets} contains datasets derived from social network interactions, crucial for tasks such as social analysis, sentiment modeling, and community detection in the Computer Science field. These datasets focus on user interactions in online platforms, which are key for understanding social dynamics and sentiment trends.
Table~\ref{tab: Other Datasets} includes datasets from domains outside the primary categories, contributing to specialized research areas, as discussed in the main text.

In addition to node-level and edge-level tasks, the datasets listed in Table~\ref{tab: Graph-level tasks datasets} are frequently used for graph-level tasks, such as graph classification, molecule classification, and graph regression. These research directions are particularly relevant to AI4Science, with applications in fields like medical research and pharmaceuticals, where graph-based models are used to study molecular structures, drug interactions, and disease modeling.

 \begin{table*}[htbp]
 \centering
\setlength{\abovecaptionskip}{0.2cm}
\caption{The statistical information of Citation Network and Wikipage Datasets. ( \textbf{.Imp} indicates the Underlying implications for nodes, edges, and classes in datasets, \textbf{Node.C} denotes for Node Classification Task,  \textbf{Edge.E} denotes for Edge Existence Task} 
\label{tab: Citation Network and Wikipage derived datasets}
\resizebox{1\textwidth}{!}{
\setlength{\tabcolsep}{0.8mm}{
\begin{tabular}{cccccccccc}
\midrule[0.3pt]
Datasets    & Nodes  & edges  & class    & Node.Imp & Edge.Imp & Class.Imp & Task & Category & Citation       \\ \midrule[0.3pt]
Cora.1      & 2,707  & 10,556  & 7   &  Scientific Publications   & co-citation       & Predefined fields &  Node.C \; Edge.E & Citation Network  & ~\cite{OMOG}   \\
Cora.2      & 2,708  & 5,278  & 7   &  Scientific Publications     & co-citation       & Predefined fields &  Node.C \; Edge.E & Citation Network   & ~\cite{he2024harnessingexplanationsllmtolminterpreter} \\
Citeseer      & 3,186  & 4,277  & 6   & Scientific Publications     & co-citation       & Predefined subjects &  Node.C \; Edge.E  & Citation Network  & ~\cite{GAGA} \\
Pubmed.1      & 19,716  & 88,648  & 2   &  Scientific Publications    & co-citation       & predefined subjects &  Node.C \; Edge.E & Citation Network   & ~\cite{GOFA} \\
Pubmed.2      & 19,716  & 44,338  & 3   &  Scientific Publications     & co-citation       & Predefined subjects &  Node.C \; Edge.E  & Citation Network  & ~\cite{he2024harnessingexplanationsllmtolminterpreter}\\
Ogbn-Arxiv      & 169,343  & 1,166,243  & 40   &  arXiv Papers     & co-citation       & Predefined subjects &  Node.C \; Edge.E  & Citation Network  & ~\cite{chen2024llagalargelanguagegraph} \\
DBLP      & 110,757  & 655,766  & 30  &  DBLP papers    & co-citation       & Predefined subjects &  Node.C \; Edge.E  & Citation Network  & ~\cite{pan2024tagexplainernarratinggraphexplanations} \\
Ogbn-Arxiv-2023      & 46,198  & 78,548  & 40   &  arXiv Papers     & co-citation       & Predefined subjects &  Node.C \; Edge.E & Citation Network  & ~\cite{zhu2024efficienttuninginferencelarge}\\
Cora.Full      & 19,793  & 126,842  & 70   &  Scientific Publications    & co-citation       & Predefined subjects &  Node.C \; Edge.E  & Citation Network  & ~\cite{zhao2024graphanyfoundationmodelnode}   \\
ACM      & 48,579  & 193,034  & 9   & Scientific Publications    & citation       & Predefined subjects &  Node.C \; Edge.E & Citation Network   & ~\cite{SKETCH}    \\\
Ogbl-citation2      & 2,927,863  & 30,387,995  & 172   &   Papers     & co-citation       & Predefined subjects &  Edge.E & Citation Network  & ~\cite{duan2023simtegfrustratinglysimpleapproach} \\\
Tape-Arxiv23      & 46,198  & 78,543  & 40   &  papers     & co-citation       & Predefined subjects &  Node.C \; Edge.E & Citation Network  & ~\cite{GAGA} \\ \midrule[0.3pt]
WikiCS     & 11,701  & 216,123  & 10   &  Wiki Page     & hyperlinks      & Predefined Page category &  Node.C \; Edge.E  & Wikipedia Network  & ~\cite{seo2024unleashingpotentialtextattributedgraphs} \\
Cornell     & 247  & 213  & 5   &  Web Page     & hyperlinks      & Predefined Page Category &  Node.C \; Edge.E & Web Page Network  & ~\cite{zhao2024graphanyfoundationmodelnode}\\
Texas     & 255  & 119  & 5   &  Web Page     & hyperlinks      & Predefined Page Category &  Node.C \; Edge.E  & Web Page Network & ~\cite{zhao2023graphtextgraphreasoningtext}  \\
Wisconsin    & 320  & 449  & 5   &  Web Page     & hyperlinks      & Predefined Page Category &  Node.C \; Edge.E & Web Page Network  & ~\cite{SCORE} \\
Actor     & 7,600  & 30,019  & 5   &  actors     &  Co-occurrence    & Common Words on Page   & Node.C & Web Page Network & ~\cite{anonymous2024edge} \\
Wiki     & 36,501  & 1,190,369  & 10   &  articles     & Hyperlink    & Predefined categories & Node.C & Wikipedia Network  & ~\cite{SKETCH} \\ 
Chameleon     & 2,277  & 36,101  & 5   &  articles     & co-linkage    & average monthly traffic   & Node.C & Wikipedia Network & ~\cite{wang2024learninggraphquantizedtokenizers}  \\ 
Squirrel     & 5,201  & 217,073  & 5   &  articles     & co-linkage    & average monthly traffic  & Node.C & Wikipedia Network & ~\cite{wang2024learninggraphquantizedtokenizers} \\
Roman Empire     & 22,662  & 65,854  & 18   &  non-unique words     & Contextual structure    & syntatic role of word & Node.C  &  Wikipedia network & ~\cite{wang2024learninggraphquantizedtokenizers} \\ \midrule[0.3pt]
FB15k    & 14,951  &   & 1345(relation types)  &  Real-world entities     &  connectivity    & relation types between entities   & KG Completion & Knowledge Graph & ~\cite{SCORE} \\
WN18    & 40,943  &   & 18(relation types)  &  words/concept    &  connectivity    & semantic relationship types   & KG Completion & Knowledge Graph & ~\cite{SCORE} \\ \midrule[0.3pt]

\end{tabular}
}}
\vspace{-0.10cm}
\end{table*}

   \begin{table*}[htbp]
\setlength{\abovecaptionskip}{0.2cm}
\caption{The statistical information of E-Commerce Datasets, including Rating Network, Review Network}
\small 
\label{tab: E-commerce Datasets}
\resizebox{1\textwidth}{!}{
\setlength{\tabcolsep}{2mm}
\begin{tabular}{cccccccccc}
\midrule[0.3pt]
Datasets    & Nodes  & Edges  & Class    & Node.Imp & Edge.Imp & Class.Imp & Task & Category & Ciation         \\ \midrule[0.3pt]
A-Computer      & 13,381   & 245,778  & 10   &  product   & co-purchase       & Predefined product categories &  Node.C \; Edge.E & Co-purchase Dataset  & ~\cite{mavromatis2023traingnnteachergraphaware}   \\
A-photo      & 7,487  & 119,043  & 8   &  product   & co-purchase       & Predefined product categories &  Node.C \; Edge.E & Co-purchase Dataset & ~\cite{mavromatis2023traingnnteachergraphaware}  \\
Amazon Computer      & 87,229  & 808,310  & 10    &  product   & co-purchase       & Predefined product categories &  Node.C \; Edge.E & Co-purchase Dataset  & ~\cite{zhang2024tagatextattributedgraphselfsupervised} \\
Amazon Photo     & 48,362  & 549,290  & 12    &  product   & co-purchase       & Predefined product categories &  Node.C \; Edge.E & Co-purchase Dataset & ~\cite{gppt}  \\ 
Ogbn-Products      & 54,025  & 74,420  & 47    &  product   & co-purchase       & product categories &  Node.C \; Edge.E & Co-purchase Dataset & ~\cite{chien2022nodefeatureextractionselfsupervised}   \\ \midrule[0.3pt]
Child     & 76,875  & 1,554,578  & 24   &  Books   & co-purchase/viewed       & Predefined categories &  Node.C \; Edge.E  & E-commerce datasets & ~\cite{OMOG}  \\
History     & 41,551  & 400,125  & 12   &  Books   & co-purchase/viewed       & Predefined categories &  Node.C \; Edge.E & E-commerce datasets  & ~\cite{pan2024tagexplainernarratinggraphexplanations}  \\
Sports-fitness    & 173,055  & 1,773,594  & 13   &  products   & co-purchase/viewed       & Products categories &  Node.C \; Edge.E & E-commerce datasets  & ~\cite{zhang2024tagatextattributedgraphselfsupervised} \\
Ele-Photo    & 48,362  & 500,928  & 12   &  products   & co-purchase/viewed       & Products categories &  Node.C \; Edge.E & E-commerce datasets & ~\cite{zhu2024efficienttuninginferencelarge}  \\
Art    & 1,615,902  & 4,898,218  & 3347   &  products   & co-viewed       & fine-grained classes &  Node.C \; Edge.E & E-commerce datasets & ~\cite{G2P2}  \\
Industrial    & 1,260,053  & 3,101,670  & 2462   &  products   & co-viewed       & fine-grained classes &  Node.C \; Edge.E & E-commerce datasets & ~\cite{G2P2}  \\
Music Instrument    & 905,453  & 2,692,734  & 1191   &  products   & co-viewed       & fine-grained classes &  Node.C \; Edge.E & E-commerce datasets  & ~\cite{zhu2024efficienttuninginferencelarge} \\
Amazon    & 50,000  & 632,802  & 7   &  textual description   & co-viewed       & Predefined categories &  Node.C & E-commerce datasets & ~\cite{SKETCH}  \\ 
AmzComp   & 13,752  & 491,722  & 10   &   product   & co-purchase       & Predefined categories &  Node.C & E-commerce datasets  & ~\cite{mavromatis2023traingnnteachergraphaware} \\ 
Amazon-Rating   & 24,492  & 1,866,100  & 5   &   product   & Co-rated       & Predefined rating levels &  Node.C & E-commerce datasets & ~\cite{GIT} \\ \midrule[0.3pt]
Goodreads-Children   & 192,036  & 734,640  &    &    User, Books   & User-book, co-author  & Book genre &  Node.C & Book System datasets & ~\cite{Link2Doc} \\
Goodreads-Crime   & 385,203  & 1,849,236  &    &   User, Books   & User-book, co-author  & Book genre &  Node.C & Book System datasets & ~\cite{Link2Doc} \\ \midrule[0.3pt]

\end{tabular}
}
\vspace{-0.10cm}
\end{table*}

    \begin{table*}[htbp]
\setlength{\abovecaptionskip}{0.2cm}
\caption{The statistical information of Social Network Datasets, including Online Community datasets, Co-authorship Network }
\footnotesize 
\label{tab: Social Network Datasets}
\resizebox{1\textwidth}{!}{
\setlength{\tabcolsep}{1mm}{
\begin{tabular}{cccccccccc}
\midrule[0.3pt]
Datasets    & Nodes  & Edges  & Class    & Node.Imp & Edge.Imp & Class.Imp & Task & Category & Citation         \\ \midrule[0.3pt]
IMDB     & 4,182   & 47,789  & 3   &  movies   & overlapped professionals       & Predefined genres &  Node.C \; Edge.E & Movie Network Dataset & ~\cite{seo2024unleashingpotentialtextattributedgraphs}    \\
Reddit     & 232,965   & 114,615,892  & 41   &  posts   & co-comments       & Community subjects &  Node.C & Online Community Dataset  & ~\cite{gppt}   \\
Pokec      & 1,632,803  & 30,662,564  & 2   &  users   & friendship       & Binary gender of users &  Node.C \; Edge.E & Social Network Dataset & ~\cite{anonymous2024gfse}  \\
LastFM Asia      & 7,624  & 55,612  & 18    &  users   & co-followship       & Predefined nationality &  Node.C & Social Network Dataset  & ~\cite{zhao2024graphanyfoundationmodelnode}  \\
Deezer      & 28,281  & 185,504  & 2    &  users   & co-followship       & Binary gender of users &  Node.C  & Social Network Dataset  & ~\cite{zhao2024graphanyfoundationmodelnode}  \\
Questions      & 48,921  & 307,080 & 2   &  users   & comments      & Activity Level &  Node.C  & Social Network Dataset & ~\cite{zhao2024graphanyfoundationmodelnode}  \\
BlogCatalog      & 5,196  & 17,981  & 6   &  bloggers   & friendship       & Predefined labels &  Node.C & Social Network Dataset  & ~\cite{anonymous2024graphfm}  \\ \midrule[0.3pt]
Co-author Physics      & 34,493  & 495,924  & 5   &  author   & co-authorship       & Most active field for authors &  Node.C  & Co-authorship Dataset & ~\cite{zhao2024graphanyfoundationmodelnode} \\
Co-author CS      & 18,333  & 163,788  & 15   &  author   & co-authorship       & Most active field for authors &  Node.C   & Co-authorship Dataset & ~\cite{zhao2024graphanyfoundationmodelnode} \\
Tokoler      & 11,758  & 1,038,000  & 2   &  worker   & collague on the project       & Bianry activity of workers &  Node.C & Crowd-sourcing Dataset   & ~\cite{zhao2024graphanyfoundationmodelnode}\\ \midrule[0.3pt]
\end{tabular}
}}
\vspace{-0.10cm}
\end{table*}

    \begin{table*}[htbp]
\setlength{\abovecaptionskip}{0.2cm}
\caption{The statistical information of other datasets for node-level and link-level tasks}
\footnotesize 
\label{tab: Other Datasets}
\resizebox{1\textwidth}{!}{
\setlength{\tabcolsep}{2mm}{
\begin{tabular}{cccccccccc}
\midrule[0.3pt]
Datasets    & Nodes  & Edges  & Class    & Node.Imp & Edge.Imp & Class.Imp & Task  & Category & Citation        \\ \midrule[0.3pt]
MIMIC-III     & 32,267   & 559,290  & 19   &  patients   & Co-interaction       & Disease categories &  Node.C & Medical Dataset & ~\cite{tan2024musegraphgraphorientedinstructiontuning}     \\
Ogbn-proteins     & 132,534   & 39,561,252  & 112   &  proteins   & correlations      & Protein Types &  Node.C & Protein Dataset & ~\cite{anonymous2024gfse}    \\
Air Brazil     & 131   & 1,074  & 4   &  airport   & commercial flights       & levels of airport activities &  Node.C & Air-Traffic Dataset & ~\cite{zhao2024graphanyfoundationmodelnode}    \\
Air EU     & 299   & 5,995  & 4   &  airport   & commercial flights       & levels of airport activities &  Node.C & Air-Traffic Dataset  & ~\cite{zhao2024graphanyfoundationmodelnode}   \\
Air US     & 1,190   & 13,599  & 4   &  airport   & commercial flights       & levels of airport activities &  Node.C & Air-Traffic Dataset  & ~\cite{zhao2024graphanyfoundationmodelnode}   \\
Minesweeper     & 10,008   & 78,812  & 2   &  cell   & conncetivity       & Binary class of mine existence &  Node.C & Grid-by-grid Dataset & ~\cite{zhao2024graphanyfoundationmodelnode}    \\ \midrule[0.3pt]

\end{tabular}
}}
\vspace{-0.10cm}
\end{table*}

    \begin{table*}[htbp]
\setlength{\abovecaptionskip}{0.2cm}
\caption{The statistical information of datasets for Graph-level Tasks. \textbf{Avg Nodes} and \textbf{Avg Edges} are denoted for the averaged number of nodes and edges in subgraphs, respectively}
\footnotesize 
\label{tab: Graph-level tasks datasets}
\resizebox{\linewidth}{33mm}{
\setlength{\tabcolsep}{7mm}{
\begin{tabular}{ccccccc}
\midrule[0.3pt]
Datasets    & \# Graph  & \# Graph Class  & \# Avg Nodes   & \# Avg Edges & Task  & Citations     \\ \midrule[0.3pt]
PROTEINS    & 1,113    & 2  & 39.06   &  72.82   & Protein Graph Classification   & ~\cite{SCORE} \\
DD    & 1,178    & 2  & 284.1   &  715.7   & Protein Graph Classification   & ~\cite{SCORE} \\
ENZYMES    & 600    & 3  & 32.63   &  62.14   & Protein Graph Classification   & ~\cite{liu2023graphpromptunifyingpretrainingdownstream} \\
COX2    & 467    & 2  & 41.22   &  43.45   & Small Molecule Classification   & ~\cite{liu2023graphpromptunifyingpretrainingdownstream} \\
BZR    & 405    & 2  & 35.75   &  38.36   & Small Molecule Classification   & ~\cite{jiang2024ragraphgeneralretrievalaugmentedgraph} \\
MUTAG    & 188    & 2  & 17.9   &  19.8   & Small Molecule Classification   & ~\cite{SCORE} \\
Mutagenicity    & 4,337    & 2  & 30.32   &  61.54   & Small Molecule Classification   & ~\cite{anonymous2024edge} \\
PCBA    & 437,929    & 128  & 25.97   &  28.11   & Molecule Binary Classification   & ~\cite{anonymous2024edge} \\
HIV    & 41,127    & 2  & 25.51   &  27.46   & Molecule Binary Classification   & ~\cite{he2024unigraphlearningunifiedcrossdomain} \\
AIDS    & 2,000    & 2  & 15.69   &  16.2   & Molecule Classification   & ~\cite{he2024unigraphlearningunifiedcrossdomain} \\
BBBP   & 2,039    & 1  & 23.9   &  51.6  & Molecule  Classification   & ~\cite{li2024finemoltexfinegrainedmoleculargraphtext} \\
Tox21   & 7,831    & 12  & 18.6   &  38.6  & Molecule  Classification   & ~\cite{li2024finemoltexfinegrainedmoleculargraphtext} \\
ToxCast   & 8,597    & 617  & 18.7   &  38.4  & Molecule  Classification   & ~\cite{li2024finemoltexfinegrainedmoleculargraphtext} \\
SIDER   & 1,427    & 27  & 33.6   &  70.7  & Molecule  Classification   & ~\cite{li2024finemoltexfinegrainedmoleculargraphtext} \\
ClinTox   & 1,484    & 2  & 26.1   &  55.5  & Molecule  Classification   & ~\cite{li2024finemoltexfinegrainedmoleculargraphtext} \\
MUV   & 93,087    & 17  & 24.2   &  52.6  & Molecule  Classification   & ~\cite{li2024finemoltexfinegrainedmoleculargraphtext} \\
BACE   & 1,513    & 1  & 34.1   &  73.7  & Molecule  Classification   & ~\cite{li2024finemoltexfinegrainedmoleculargraphtext} \\
NCI1   & 4,110    & 2  & 29.87   &  64.6   & Molecule Binary Classification   & ~\cite{SCORE} \\
Peptides-func   & 15,535    & 10  & 150.94   &  153.65   & Molecule Binary Classification   & ~\cite{anonymous2024gfse} \\
Peptides-struc   & 15,535    & 11  & 150.94   &  153.65   & Regression   & ~\cite{anonymous2024gfse} \\
Zinc   & 294,456    & 1  & 23.2   &  49.8   & Regression   & ~\cite{anonymous2024gfse} \\
CIFAR   & 60,000    & 1  & 117.6   &  941.2   & classification   & ~\cite{anonymous2024gfse} \\
ChEMBL   & 365,065   & 10  & 25.9   &  55.9   & classification   & ~\cite{he2024unigraphlearningunifiedcrossdomain} \\
IMDB-Binary    & 1,000    & 2  & 19.8   &  96.53   & Social Network Classification   & ~\cite{SCORE} \\
COLLAB    & 5,000    & 3  & 74.5   &  2457.8   & Social Network Classification   & ~\cite{SCORE} \\
Reddit-B    & 2,000    & 2  & 429.63   &  497.75   & Binary Classification   & ~\cite{anonymous2024graphprop} \\
Reddit-M5K    & 4,999    & 5  & 508.52   &  594.87   & Classification   & ~\cite{anonymous2024graphprop} \\ \midrule[0.3pt]
\end{tabular}
}}
\vspace{-0.10cm}
\end{table*}

\subsection{Detailed Datasets Description:}

    For readers who intend to select the combinations of diversified datasets for research purposes, we provide detailed descriptions for the datasets. 
    By clearly outlining where each dataset comes from and the specific context in which it is applied, we give readers a deeper understanding of how these datasets are utilized in different research settings. 
    This transparency allows readers to assess the relevance and suitability of each dataset for their own work, ensuring they can make informed decisions when selecting data for their own studies. 
    The details are as follows:

\textbf{MUTAG}~\cite{debnath1991_MUTAG} is a widely used bioinformatics dataset consisting of 188 graphs, each representing a nitro compound. 
    The nodes are labeled with one of 7 distinct node labels.
    The primary objective of this dataset is to classify each graph to determine whether the corresponding compound is mutagenic, specifically distinguishing between aromatic and heteroaromatic compounds.

    \textbf{BZR}~\cite{sutherland2003_COX2_BZR_DHFR} is a bioinformatics dataset used for compound activity prediction, with a primary focus on a collection of benzimidazole compounds. 
    The dataset is designed to indicate the concentration of each compound necessary to inhibit the activity of specific biomolecules, providing valuable insights into the effectiveness of these compounds in biological processes.

    \textbf{COX2}~\cite{sutherland2003_COX2_BZR_DHFR} is a dataset centered on Cyclooxygenase-2, an enzyme that plays a critical role in inflammation and pain mechanisms. 
    The dataset is utilized to classify various compounds and predict their potential inhibition potency against the COX-2 enzyme, which is a key target in drug development for anti-inflammatory therapies.

    \textbf{AIDS}~\cite{riesen2008_AIDS} is a graph dataset comprising 2000 graphs, each representing molecular compounds derived from the AIDS Antiviral Screen Database of Active Compounds. 
    The dataset includes a total of 4395 chemical compounds, categorized into three classes: 423 compounds belonging to class CA, 1081 to class CM, and the remaining compounds to class CI. 
    This dataset is widely used for tasks involving molecular classification and drug discovery research.

    \textbf{NCI1}~\cite{wale2008_NCI1} is a bioinformatics dataset comprising 4,110 graphs representing chemical compounds. 
    It contains data published by the National Cancer Institute (NCI). 
    Each node is assigned one of 37 discrete node labels. 
    The graph classification label is determined by NCI anti-cancer screens assessing the ability to suppress or inhibit the growth of a panel of human tumor cell lines.

    \textbf{ENZYMES}~\cite{borgwardt2005_ENZYMES} is a comprehensive dataset containing 600 protein tertiary structures, meticulously curated from the BRENDA enzyme database. 
    Within the ENZYMES dataset, researchers can explore the intricate structures of six unique enzymes, providing a rich resource for computational analysis and machine learning applications.

    \textbf{DD}~\cite{dobson2003_DD} is a bioinformatics dataset composed of 1,178 graph structures representing proteins. 
    In these graphs, nodes correspond to amino acids, and edges connect nodes that are within 6 Angstroms of each other, reflecting the spatial proximity of amino acids within the protein structure. 
    The primary task associated with this dataset is a binary classification to differentiate between enzymes and non-enzymes, making it a valuable resource for studies in protein function prediction and structural bioinformatics.

    \textbf{PROTEINS}~\cite{helma2001_PTC_PROTEINS} is a bioinformatics dataset comprising 1,113 structured proteins. 
    Nodes in these graph-based proteins denote secondary structure elements and are assigned discrete node labels indicating whether they represent a helix, sheet, or turn. 
    Edges indicate adjacency along the amino-acid sequence or in space between two nodes. 
    The objective is to predict the protein function.

    \textbf{COLLAB}~\cite{leskovec2005graphs_COLLAB} is a scientific collaboration dataset consisting of 5,000 ego networks represented as graphs. 
    This dataset is compiled from three public collaboration datasets. 
    Each ego network comprises researchers from various fields and is labeled according to the corresponding field, namely High Energy Physics, Condensed Matter Physics, and Astrophysics.

    \textbf{IMDB-BINARY}~\cite{yanardag2015_IMDB_B_M} is a movie collaboration dataset comprising 1,000 graphs representing ego networks for actors and actresses.
    Derived from collaboration graphs within the Action and Romance genres, each graph features nodes representing actors/actresses and edges denoting their collaboration in the same movie. 
    Graphs are labeled according to the corresponding genre, and the objective is to classify the genre for each graph.

    \textbf{IMDB-MULTI}~\cite{yanardag2015_IMDB_B_M} is the multi-class extension of the IMDB-BINARY dataset, comprising 1,500 ego-networks. 
    It includes three additional movie genres: Comedy, Romance, and Sci-Fi, making it suitable for multi-class classification tasks. 
    This dataset is commonly used to evaluate the performance of graph-level algorithms.

    
    \textbf{Cora}, \textbf{CiteSeer}, and \textbf{PubMed}~\cite{Yang16cora} are widely used citation network datasets, where nodes represent papers and edges denote citation relationships.
    Node features are word vectors, indicating the presence or absence of specific words in each paper. 
    These datasets are frequently used for node classification.

    \textbf{ogbn-arxiv}~\cite{hu2020ogb} is a widely used citation graph indexed by Microsoft Academic Graph (MAG)~\cite{wang2020microsoft_MAG}, especially for the large-scale graph learning. 
    Each paper in the dataset is represented by the average of the word embeddings derived from its title and abstract. 
    These word embeddings are generated using the skip-gram model, which captures semantic relationships between words based on their context within the text.
    This dataset is widely used for graph-based learning tasks, such as node classification.

    \textbf{Amazon Photo} and \textbf{Amazon Computers}~\cite{shchur2018amazon_datasets} are subsets of the Amazon co-purchase graph, where nodes represent individual products, and edges signify that two products are frequently bought together. 
    The node features for these datasets are derived from product reviews, represented as bag-of-words vectors, capturing the textual information associated with each item. 
    These datasets are commonly used for graph-based downstream tasks such as node classification in graph-based recommendation systems.

    \textbf{ogbn-products}~\cite{hu2020ogb} is a co-purchasing network where nodes represent products and edges indicate frequent co-purchases. 
    The node features are derived from bag-of-words representations of product descriptions. 
    Due to its extensive size and complex structure, this dataset is particularly well-suited for large-scale graph learning applications, making it an ideal benchmark for evaluating the scalability and performance of graph-based algorithms.

    \textbf{Coauthor CS} and \textbf{Coauthor Physics}~\cite{shchur2018amazon_datasets} are co-authorship graphs derived from the MAG~\cite{wang2020microsoft_MAG}. 
    In these graphs, nodes represent individual authors, edges denote co-authorship relationships between them, and node features are constructed from the keywords of the authors' publications. 
    The labels assigned to the nodes indicate the specific research fields in which the authors are active. 
    These datasets are commonly used for evaluating graph-based methods, particularly in the context of node classification.

    \textbf{Chameleon} and \textbf{Squirrel}~\cite{wang2024learninggraphquantizedtokenizers} are two page-page networks extracted from specific topics within Wikipedia. 
    In these datasets, nodes represent web pages, while edges signify mutual links between pages. 
    Node features are derived from several informative nouns found on Wikipedia. 
    They categorize the nodes into five groups based on the average monthly web page traffic.

    \textbf{Actor}~\cite{anonymous2024edge} is an actor co-occurrence network where nodes represent actors, and edges indicate their co-appearance on Wikipedia pages. 
    Node features are bag-of-words vectors derived from these pages, and actors are categorized into five groups based on the terms found in their respective Wikipedia entries. 
    This dataset is commonly used for graph-based tasks like node classification.

    \textbf{Minesweeper}~\cite{zhao2024graphanyfoundationmodelnode} draws inspiration from the Minesweeper game and stands as the synthetic dataset. 
    The graph is a regular 100x100 grid, where each node (cell) is linked to its eight neighboring nodes (excluding nodes at the grid's edge, which have fewer neighbors). 
    Twenty percent of the nodes are randomly designated as mines. 
    The objective is to predict which nodes conceal mines. 
    Node features consist of one-hot-encoded counts of neighboring mines. 
    However, for a randomly chosen 50\% of the nodes, the features are undisclosed, indicated by a distinct binary feature.

    \textbf{Tolokers}~\cite{zhao2024graphanyfoundationmodelnode} is derived from the crowdsourcing platform~\cite{Tolokers_original}.
    Nodes correspond to workers who have engaged in at least one of the 13 selected projects. 
    An edge connects two workers if they have collaborated on the same task. 
    The objective is to predict which workers have been banned in one of the projects.

    \textbf{Roman-empire}~\cite{wang2024learninggraphquantizedtokenizers} is based on the Roman Empire article from the English Wikipedia~\cite{lhoest2021empire_original}, each node corresponds to a non-unique word in the text, mirroring the article's length. 
    Nodes are connected by an edge if the words either follow each other in the text or are linked in the sentence's dependency tree. 
    Thus, the graph represents a chain graph with additional connections.

    \textbf{Amazon-ratings}~\cite{GIT} is derived from the co-purchasing network and its metadata available in the SNAP~\cite{leskovec2014rating_original}. 
    Nodes are items and edges connect items frequently bought together. 
    The task is predicting the average rating given by reviewers, categorized into five classes. 
    Node features are based on the FastText embeddings~\cite{grave2018fast_word_embedding} of words in the product description.
    To manage graph size, only the largest connected component of the 5-core is considered.

    \textbf{Questions}~\cite{zhao2024graphanyfoundationmodelnode} is derived from data collected from the question-answering platform Yandex Q. 
    In this dataset, nodes represent users, and an edge exists between two nodes if one user answers another user's question within a one-year timeframe (from September 2021 to August 2022). 
    The objective is to predict which users remained active on the website (i.e., were not deleted or blocked) by the end of the specified period. 
    For node features, it utilizes the average FastText embeddings for words found in the user descriptions. 
    
    \textbf{Pokec}~\cite{anonymous2024gfse} is a social network, where nodes are users, and edges are friendships. The task is to predict the gender. 

    \textbf{Tape-Arxiv23}~\cite{he2024harnessingexplanationsllmtolminterpreter} is a directed graph of citation networks among computer science arXiv papers from 2023 onwards. Each node represents a paper, and the edges show citation links. The task is to predict the 40 subject areas of these papers, such as cs.AI, cs.LG, and cs.OS, based on author and moderator labels.

    \textbf{Ele-Computer/Photos}~\cite{zhu2024efficienttuninginferencelarge} both are extracted from the Amazon-Electronics dataset. Ele-Computers consists of items with the second-level label "Computers", while Ele-Photo consists of items with the second-level label "Photo". The two datasets are extracted from the updated 2018 Amazon Computer and Amazon Photo datasets. The nodes in the dataset are electronics-related products, and the edge between the two products means that they are frequently co-purchased or co-viewed. The label of each dataset is the three-level label of the electronics products. We adopt user reviews on the item as its text attribute. Since the item has multiple reviews, we mainly adopt the review with the highest number of votes. For some items lacking highly voted reviews, we randomly adopt a user review as the text attribute. The task of the two datasets is to classify electronics products into 10 and 12 categories, respectively.

    \textbf{Goodreads}~\cite{Link2Doc} is collected by the popular social media platform Goodreads, which allows users to track, review, and rate books. The datasets primarily include information on books, users, authors, reviews, ratings, and shelves (e.g., "to-read", "currently-reading", "read"). They provide rich metadata about books, such as titles, genres, descriptions, and publication details, as well as user interactions, including ratings, reviews, and social connections.

    \textbf{Art, Industrial, and Music Instrument}~\cite{G2P2} are three Amazon review datasets, respectively from three broad areas, namely, arts, crafts, and sewing(Art), industrial and scientific(Industrial), and musical instruments(M.I.). The description of each product is deemed a text document, whereas the reviews of a user are combined into one document to reflect the user’s preferences. If a user has reviewed a product, a link is constructed between them. The product subcategories within a broad area present the classes, which are fine-grained and may involve thousands of classes with subtle differences. The classification is only performed on product descriptions, whereas the user reviews only serve to enrich the text semantics.

    \textbf{MIMIC-III}~\cite{tan2024musegraphgraphorientedinstructiontuning} s a graph of diseases, patients, and visits, where nodes and relations are extracted from clinical records [26]. Diseases are classified into 19 categories according to ICD-9-CM.

    \textbf{Deezer}~\cite{zhao2024graphanyfoundationmodelnode} is A social network of Deezer users that was collected from the public API in March 2020. Nodes are Deezer users from European countries and edges are mutual follower relationships between them. The vertex features are extracted based on the artists liked by the users. The task related to the graph is binary node classification - one has to predict the gender of users. This target feature was derived from the name field for each user.

    \textbf{LastFM Asia}~\cite{zhao2024graphanyfoundationmodelnode} is an online social network of people who use the online music streaming site LastFM and live in Asia. The links represent reciprocal follower relationships and the vertex features describe the list of musicians liked by the users. The machine learning task is the prediction of nationality for the users of the site.

    \textbf{BlogCatalog}~\cite{anonymous2024graphfm} is a social network with bloggers and their social relationships. Node features are constructed by the keywords of user profiles, and the labels are the topic categories provided by the authors. 

    \textbf{Reddit-Multi (5k)}~\cite{anonymous2024graphprop} is generated in a similar way to Reddit-Binary. The difference is that there are World News, Videos, Advice Animals, and Mildly Interesting. Graphs are labeled with their corresponding subreddits.

    \textbf{Reddit-Binary}~\cite{anonymous2024graphprop} consists of graphs corresponding to online discussions on Reddit. In each graph, nodes represent users, and there is an edge between them if at least one of them responds to the other’s comment. There are four popular subreddits, namely, IAmA, AskReddit, TrollXChromosomes, and atheism. IAmA and AskReddit are two question/answer-based subreddits, and TrollXChromosomes and atheism are two discussion-based subreddits. A graph is labeled according to whether it belongs to a question/answer-based community or a discussion-based community.

    \textbf{FB-15k}~\cite{SCORE} was introduced in "Translating Embeddings for Modeling Multi-relational Data". It is a subset of Freebase which contains about 14,951 entities with 1,345 different relations. When creating the dataset, a reverse edge with reversed relation types is created for each edge by default.

    \textbf{WN18RR}~\cite{SCORE} was introduced in "Translating Embeddings for Modeling Multi-relational Data". It included the full 18 relations scraped from WordNet for roughly 41,000 synsets. When creating the dataset, a reverse edge with reversed relation types is created for each edge by default.

\vspace{-0.3cm}

\section{Why GNN Needs LLM: The Role of Models}

In Graph Neural Networks (GNNs), downstream tasks are often closely tied to specific data domains, leading to research that focuses on domain-specific tasks with tailored datasets. For example, a GNN trained on a citation network for node classification is optimized for task-specific knowledge, but this results in significant performance degradation when applied to other domains. Such deficits reveal GNNs' heavy reliance on the alignment between task and data domains.


While maintaining consistency between data and task domains during both training and inference can ensure reliable performance, this assumption fails to meet real-world demands, where graph data and task requirements often vary widely. To address this, integrating large pre-trained models, such as Large Language Models (LLMs), as collaborative modules with GNNs offers an effective solution by improving generalizability across diverse domains.


By leveraging the strong representation and generalization capabilities of LLMs during training, collaborative training and inference between GNNs and LLMs can be achieved. Using textual forms of node or edge attributes, this approach enhances cross-domain generalization in graph-based models. Hence, it eliminates the need for repetitive domain-specific retraining for adapting to varying scenarios, overcoming the limitations of existing models and gradually advancing towards the development of a Graph Foundation Model.

Building on these insights, we categorize existing research into five perspectives to explore various collaborative approaches between GNNs and LLMs and their applications in graph learning tasks. Specifically, we aim to separate research based on their model architecture designs and describe the roles of LLMs and GNNs have contributed to the training process. The comprehensive summaries are provided as follows :
\begin{itemize}

\vspace{-0.05cm}

    \item \textbf{GNN and LLM as independent collaborative modules}: GNNs and LLMs operate as two independent components, to jointly solve graph-related tasks. The representative research papers include the following: ConGraT\cite{brannon2024congratselfsupervisedcontrastivepretraining}, SimTeG\cite{duan2023simtegfrustratinglysimpleapproach}, AUGGLM\cite{xu2024languagemodelsgraphlearners}, FineMolTex\cite{li2024finemoltexfinegrainedmoleculargraphtext}, GraphToken\cite{perozzi2024letgraphtalkingencoding}, Link2Doc\cite{Link2Doc}, TAGA\cite{zhang2024tagatextattributedgraphselfsupervised}, UniGraph\cite{he2024unigraphlearningunifiedcrossdomain}, CIKM-KD\cite{pan2024distillinglargelanguagemodels}, CurGL\cite{anonymous2024curriculum}, Grenade\cite{li2023grenadegraphcentriclanguagemodel}, THLM\cite{THLM}, RoSE\cite{seo2024unleashingpotentialtextattributedgraphs}, GraphBridge\cite{anonymous2024graphbridge}, GAGA\cite{GAGA}, GIANT\cite{chien2022nodefeatureextractionselfsupervised}, GLEM\cite{zhao2023learninglargescaletextattributedgraphs}, LLM-GNN\cite{chen2024labelfreenodeclassificationgraphs}, TAPE\cite{he2024harnessingexplanationsllmtolminterpreter}, LMGJoint\cite{LMGJoint}, OMOG\cite{OMOG}, G2P2\cite{G2P2}, TouchUp-G\cite{zhu2023touchupgimprovingfeaturerepresentation}, SKETCH\cite{SKETCH}, GraphTranslator\cite{zhang2024graphtranslatoraligninggraphmodel}, Mol-MVMoE\cite{MolMvMOE}.

    \item \textbf{GNN-enhanced LLM (Learnable)}: GNNs serve as auxiliary modules for LLMs, leveraging their powerful graph modeling capabilities to provide high-level graph feature representations, thereby enhancing LLM performance in graph-related tasks. The research work reflecting the current state of the field mainly includes: DGTL\cite{qin2024disentangledrepresentationlearninglarge}, GOFA\cite{xia2024anygraphgraphfoundationmodel}, HIGHT\cite{chen2024highthierarchicalgraphtokenization}, GL-Fusion\cite{anonymous2024glfusion}, LLaGA\cite{chen2024llagalargelanguagegraph}, ENGINE\cite{zhu2024efficienttuninginferencelarge}, HiGPT\cite{tang2024higptheterogeneousgraphlanguage}, ZeroG\cite{li2024zeroginvestigatingcrossdatasetzeroshot}, WalkLM\cite{WalkLM}, GQT\cite{wang2024learninggraphquantizedtokenizers}, SheaFormer\cite{SheaFormer}, PromptGFM\cite{PromptGFM}, GraphGPT\cite{tang2024graphgptgraphinstructiontuning}, GraphAdapter\cite{huang2024gnngoodadapterllms}, GraphPrompter\cite{graphprompter}, GaLM\cite{xie2023graphawarelanguagemodelpretraining}.

    \item \textbf{LLM-enhanced GNN (Learnable)}: 
    LLMs, with their strengths in textual representation, extract graph descriptions to provide high-quality features, thereby enhancing the learning capacity and performance of GNNs. 
    These kinds of studies include:
    PATTON\cite{jin2023pattonlanguagemodelpretraining}, 
GraphLLM\cite{chai2023graphllmboostinggraphreasoning}, 
AnyGraph\cite{xia2024anygraphgraphfoundationmodel}, 
GraphFM\cite{anonymous2024graphfm}, 
PGT\cite{pgt}, 
SCORE\cite{SCORE}, 
OpenGraph\cite{xia2024opengraphopengraphfoundation}, 
GFSE\cite{anonymous2024gfse}, 
OFA\cite{OFA}, 
GAugLLM\cite{fang2024gaugllmimprovinggraphcontrastive}, 
GRAD\cite{mavromatis2023traingnnteachergraphaware}, 
GraphAlign\cite{hou2024graphalignpretraininggraphneural}, 
GraphProp\cite{anonymous2024graphprop}, OMOG\cite{OMOG}.
    
    \item \textbf{GNN-only}: This approach focuses on the capabilities of GNNs themselves, optimizing the modeling of graph data to achieve specific task objectives. These papers are the following: GraphAny \cite{zhao2024graphanyfoundationmodelnode}, GraphFM \cite{anonymous2024graphfm}, DP-GPL \cite{anonymous2024dpgpl}, EdgePromp \cite{anonymous2024edge}, GraphBridge (ICLR) \cite{anonymous2024graphbridge}, All in One \cite{AllInOne}, GPF \cite{fang2023universal}, PRODIGY \cite{huang2023prodigy}, GraphPrompt \cite{liu2023graphpromptunifyingpretrainingdownstream}, MultiGPrompt \cite{yu2024multigpromptmultitaskpretrainingprompting}, RAGraph \cite{jiang2024ragraphgeneralretrievalaugmentedgraph}, GPPT \cite{gppt}, GIT\cite{GIT}.

    \item \textbf{LLM-only}: This method relies on LLMs to process graph data with textual attributes, exploring their potential in purely text-based graph tasks. The main works in this category include: InstructGraph \cite{wang2024instructgraphboostinglargelanguage}, LPNL \cite{bi2024lpnlscalablelinkprediction}, GPT4Graph \cite{guo2023gpt4graphlargelanguagemodels}, GraphText \cite{zhao2023graphtextgraphreasoningtext}, GraphAgent-Reasoner \cite{hu2024scalableaccurategraphreasoning}, TAGExplainer \cite{pan2024tagexplainernarratinggraphexplanations}, InstructGLM \cite{Instruct-GLM}, Talk like a Graph \cite{fatemi2023talklikegraphencoding}, NLGraph \cite{NLGraph}, MuseGraph \cite{tan2024musegraphgraphorientedinstructiontuning}.

\end{itemize}

\vspace{-0.1cm}

    Among these five perspectives, each method provides a unique solution to different problems in graph learning, laying a solid theoretical and practical foundation for the implementation of the Graph Foundation Model.
    In these methods, GNN and LLM as independent collaborative modules, GNN-only, and LLM-only are limited by the design of their model architectures, preventing them from fully utilizing the GNN's graph structure modeling capabilities and the LLM's natural language understanding abilities.
    In contrast, learnable methods such as GNN-enhanced LLM and LLM-enhanced GNN can effectively integrate structured graph data and unstructured text data through adaptive integration and joint training, enabling the model to learn richer feature representations and further improve performance.

\section{Overview of Included Approaches:}

By providing succinct yet intuitive summaries of the approaches included in this survey below, we aim to offer several key advantages. First, these summaries enhance accessibility by presenting complex methodologies in a simplified and digestible format, enabling both experts and newcomers to quickly grasp the core concepts without needing to dive into intricate technical details. This approach effectively bridges the gap between theoretical advancements and practical applications, focusing on the most important aspects of each method. Additionally, these concise summaries allow for efficient comparison across different approaches, making it easier to identify their strengths, weaknesses, and areas for further development. Ultimately, by distilling the essence of each approach, we hope to promote a deeper understanding of the research landscape and inspire the exploration of innovative solutions, contributing to future advancements in the field. The descriptions are as follows: 

\vspace{0.2cm}

\textbf{LPNL}~\cite{bi2024lpnlscalablelinkprediction} aims to utilize LLMs for learning tasks on large-scale heterogeneous graphs, designed specifically for link prediction tasks. Its core is the novel query prompt template, which is capable of extracting key information for the task by handling the enormous heterogeneous graph and limiting within the token length constraints.

\textbf{RoSE}~\cite{seo2024unleashingpotentialtextattributedgraphs} aims to decompose the heterogeneous semantic relations from existing edges on TAGs, which are treated as the single relationship, in a fully automated manner. Specifically, it uses the LLM-based generator and discriminator to differentiate various types of relations and then categorize each edge into corresponding relations by learning the in-context information between connected nodes via LLM-based decomposer. 

\textbf{TAGA}~\cite{zhang2024tagatextattributedgraphselfsupervised} is a self-supervised learning framework that jointly encodes the textual semantics and topological information of text nodes to generate a unified and comprehensive representation of TAGs. This design allows TAGA to effectively capture both the rich semantic context embedded in the textual descriptions and the structural relationships inherent in the graph topology. By leveraging this dual encoding approach, TAGA enhances the ability to learn more accurate and informative graph representations.

\textbf{CIKM-KD}~\cite{pan2024distillinglargelanguagemodels} proposes a novel knowledge distillation-inspired TAG learning framework, in which the Interpreter model trained on LLMs acquires rich rationale in text-level, structure-level, and message levels for enhancing its predictive capability and then aligns with the student model to enhance it perform high-quality inference without relying on the participation of LLMs.

\textbf{CurGL}~\cite{anonymous2024curriculum} aims at enhancing the alignment between GNNs and LLMs by balancing the node-based learning difficulty for TAG learning. In this paper, the authors explore the differentiated difficulty in nodes' classification especially when it requires learning from complex text and structural information for nodes in TAG. CurGL proposes the class-based node selection strategy to balance the training process while progressively including more nodes to align GNNs and LLMs with balanced textual and structural information.

\textbf{DP-GPL}~\cite{anonymous2024dpgpl} investigates the privacy leakage issue when using the graph prompt learning for sensitive tasks and further explores the deficit of the standard privacy method, DP-SGD, for it fails to provide privacy protection due to the small number of sensitive data points used to learn prompts. Based on this, the author proposes two privacy-preserving algorithms, DP-GPL and DP-GPL+W for generating a graph prompt with differential privacy guarantees.

\textbf{GRAD}~\cite{mavromatis2023traingnnteachergraphaware} addresses the scalability issues for effectively learning node representations on textual graphs. Specifically, GRAD applies the knowledge distillation process to encode graph structural information of GNN models, which function as the teacher model, into graph-free LM. Simultaneously, GNN models can benefit from learning the rich textual information on unlabeled nodes to improve their overall performance. 

\textbf{GraphBridge}~\cite{anonymous2024graphbridge} emphasizes further leveraging the strengths of both LLMs and GNNs by encoding the structural insights(contextual) into the semantic analysis. To enhance its scalability, it also develops a graph-aware token reduction module, which selectively preserves the most critical token for representation learning and allows more room for contextual text.

\textbf{GIANT}~\cite{chien2022nodefeatureextractionselfsupervised} develops a topology-aware self-supervised learning framework that enjoys the scalability advantages. The core of its effectiveness builds upon the numerical node feature extraction technique terms eXtreme to fine-tune the LM output based on the graph information, which is exemplified as the binary multi-labels neighborhood indicator.

\textbf{GLEM}~\cite{zhao2023learninglargescaletextattributedgraphs} improves computational efficiency and enables the scalable learning of TAGs on large datasets. By leveraging the variational Expectation-Maximization (EM) framework, GLEM effectively optimizes both GNNs and LMs in a unified manner. This simultaneous enhancement allows GLEM to better capture the complex interactions between textual and structural information within nodes, significantly improving performance, particularly in node classification tasks.

\textbf{TAPE}~\cite{he2024harnessingexplanationsllmtolminterpreter} offers a unique feature extraction pipeline termed LLM-to-LM interpreter that utilizes the textual modeling ability of LLMs to couple with GNNs' capability for the node classification. It first designs the prompt for LLMs to conduct the zero-shot classification and simultaneously requests the textual explanation for deliberating its decision-making process, which is then fed to the interpreter to transform into an information feature for GNNs to process.

\textbf{ENGINE}~\cite{zhu2024efficienttuninginferencelarge} enhances the memory usage and reduces the training complexity for TAG learning, ENGINE integrates a tunable ladder(termed G-Ladder) at each layer of LLMs, which integrates structural information through message passing mechanism. G-Ladder only requires updating a small subset of parameters which dramatically reduces the memory consumption. 

\textbf{GPPT}~\cite{gppt} reduces the fine-tuning cost for adopting the pre-trained models to target downstream tasks under the context of transfer learning for GNNs, GPPT proposes the graph prompting function to reformulate the node classification task (target downstream task) as the same as the link prediction task of the pre-trained GNN models.

\textbf{GAUgLLM}~\cite{fang2024gaugllmimprovinggraphcontrastive} improves the adaptability of graph contrastive learning in TAG by preserving the textual semantics while modifying raw text into numerical feature space. Furthermore, it introduces the edge modifier to complement the connectivity between nodes by leveraging the common structural and textual information. 

\textbf{HiGPT}~\cite{tang2024higptheterogeneousgraphlanguage} To break the constraints of generalizability to diverse downstream tasks for heterogeneous graph learning, HiGPT integrates the heterogeneous graph tokenizer that captures semantic relationships in different heterogeneous graphs. Additionally, the inclusion of a large corpus of heterogeneity-aware graph instruction enhances HiGPT's discriminative ability to comprehend complex heterogeneous relations and graph tokens. 

\textbf{ZeroG}~\cite{li2024zeroginvestigatingcrossdatasetzeroshot} proposes a new framework facilitating a more efficient transfer learning paradigm. It leverages the language model to encode both note attributes and class semantics for ensuring the consistency of feature dimensions across datasets. Additionally, the prompt-based sampling module enriches the node and structural information for extracted subgraphs during the training process.

\textbf{LMGJoint}~\cite{LMGJoint} offers an extensive benchmark work that has been deliberated for evaluating the link prediction performance of GCNs and the Pre-trained Language Model (PLM). Based on its investigation, the authors propose a fine-tuning architecture that can be easily integrated with any existing graph-based model and PLM-based text encoder to improve its memory and computing efficiency. 

\textbf{GPF}~\cite{fang2023universal} is a universal prompt-based tuning method that is applicable under any pre-training strategies. Instead of relying on designing the specialized prompt function for each pre-training strategy, GPF operates on the input graph's feature space to obtain the adaptive prompted graph for the downstream task. 

\textbf{SheaFormer}~\cite{SheaFormer}: To address the limitations of existing graph models in capturing complex node relationships, SheaFormer enhances the integration of textual and graph data by incorporating rich edge vectors. Specifically, it aggregates both node representations and edge vectors during message-passing to update node features, avoiding the need for fine-tuning LLMs on text-attributed graphs.

\textbf{G2P2}~\cite{G2P2} aims to address the label scarcity issues in graph-text model learning. Specifically, it proposes three graph interaction-based contrastive strategies to align text and graph representations. It also uses either handcrafted or automatic prompts on a pre-trained graph-text model to enhance the prompt-tuning efficiency. 

\textbf{SKETCH}~\cite{SKETCH} aims to directly adapt LMs for TAG task. It retrieves neighboring nodes for the target node based on both structural and text similarity. Both similarities are aggregated and the resulting outcomes carrying enriched text attributes are used as input for LM predictors. 

\textbf{GraphPrompt}~\cite{liu2023graphpromptunifyingpretrainingdownstream} unifies templates of objectives for both pre-training and downstream tasks and employs the learnable prompt for exploring the most relevant information from the pre-trained models to meet the downstream task, thus expanding pre-trained model's generalizability. 

\textbf{GraphAdapter}~\cite{huang2024gnngoodadapterllms}: To model TAG with large-scale LLMs, GraphAdapter proposes using GNN as the efficient adapter in the process. Specifically, treating GNNs as the tuning parameter employed at the final layer of the transformer can help reduce the distribution discrepancy between the pre-trained corpus and target TAG data. The method also performs the mean-pooling on the predicted logits from both the GNN adapter and LLMs for mutually optimizing their next-word prediction. Once the GraphAdapter is pre-trained, it can be fine-tuned for multiple downstream tasks.

\textbf{AugGLM}~\cite{xu2024languagemodelsgraphlearners}introduces a novel approach that enables off-the-shelf LMs to perform comparably to state-of-the-art GNNs on node classification tasks without modifying the LMs' architecture. The approach leverages two augmentation strategies: enriching Language model inputs with topological and semantic retrieval for richer context, and guiding classification through a lightweight GNN classifier to prune class candidates, resulting in LMs that outperform specialized node classifiers and demonstrating greater versatility across diverse datasets.

\textbf{DGTL}~\cite{qin2024disentangledrepresentationlearninglarge} enhances the reasoning and prediction capabilities of LLMs for TAGs by incorporating graph structure information through disentangled graph neural network layers. By operating with frozen pre-trained LLMs, DGTL reduces computational costs and increases flexibility, achieving superior performance over existing baselines and providing natural language explanations for predictions, thereby improving model interpretability.

\textbf{GQT}~\cite{wang2024learninggraphquantizedtokenizers} decouples tokenizer training from Transformer training using multi-task self-supervised learning, yielding robust and generalizable graph tokens. By utilizing residual vector quantization for hierarchical discrete tokens, GQT significantly reduces memory requirements and enhances generalization, enabling Transformer models to achieve state-of-the-art performance, including large-scale homophilic and heterophilic datasets.

\textbf{GraphAny}~\cite{zhao2024graphanyfoundationmodelnode} address challenges in graph-structured data, especially when inferring on new graphs with different feature and label spaces. GraphAny is a foundational architecture for inductive node classification that models inference as an analytical solution to a LinearGNN. GraphAny ensures generalization by learning attention scores for nodes and fusing predictions from multiple LinearGNNs based on entropy-normalized distance features, allowing it to adapt to new graph structures.

\textbf{GRAPHFM}~\cite{anonymous2024graphfm} address the scalability and generalizability limitations of GNNs that are typically trained on individual datasets. GRAPHFM is a scalable multi-graph pretraining approach. It uses a Perceiver-based encoder to compress domain-specific features into a shared latent space, enabling the model to generalize across diverse graph datasets, and demonstrating improved performance across multiple node classification tasks by pretraining on a variety of real and synthetic graphs.

\textbf{GraphProp}~\cite{anonymous2024graphprop} focuses on training Graph Foundation Models for graph-level tasks like protein classification, highlighting the importance of capturing consistent cross-domain information through graph structures rather than node features or graph labels. GraphProp emphasizes structural generalization by first training a structural Graph Foundation Model to predict graph properties, then using its structural representation for in-context learning with domain-specific features and labels, supported by data augmentation to enhance cross-domain generalization and address data scarcity.

\textbf{PATTON}~\cite{jin2023pattonlanguagemodelpretraining} enhances the generalizability of pre-training methods on Text-rich networks for various downstream tasks. PATTON leverages both textual information and network structure information within a Text-rich Network to optimize learning for tokens and documents. 

\textbf{FineMolTex}~\cite{li2024finemoltexfinegrainedmoleculargraphtext} better learns the motif graph and rectifies prior studies' neglect of the motif's knowledge. FineMolTex modifies the existing research focus to learn fine-grained motif-level knowledge while only conducting the coarse-grained learning objective on molecule graphs. 

\textbf{HIGHT}~\cite{chen2024highthierarchicalgraphtokenization} introduce a hierarchical graph tokenizer to capture high-order hierarchical structural information to improve LLM's perception of the graph. Additionally, it adopted an enriched fine-tuning dataset with hierarchical graph information to improve the graph-language alignment. 

\textbf{Instruct-GLM}~\cite{Instruct-GLM} motivates to explore LLMs' potential in graph learning to replace GNNs as the foundational model. It uses natural language prompts to describe multi-scale geometric graph structures and instruction-finetunes an LLM, enabling generative graph learning for graph-related tasks. 

\textbf{Grenade}~\cite{li2023grenadegraphcentriclanguagemodel} aims to advance self-supervised learning to improve representation learning for TAG. It combines a pre-trained language model (PLM) and GNNs encoder and leverages two algorithms, which is for enhance the structure-aware representation and mutual reinforcing learning between PLM and GNNs. 

\textbf{THLM}~\cite{THLM} aims to capture topological and heterogeneous information more effectively for Text-Attributed Heterogeneous Graphs. It proposes a new pre-training framework. The proposed framework introduces a topology-aware pre-training task using a context graph and a joint LM-GNN optimization, along with a text augmentation strategy to enrich text-deficient nodes by leveraging their neighbors’ textual information.

\textbf{GAGA}~\cite{GAGA} aims to reduce the annotation cost in collaborative training between GNNs and LLMs for TAG. Specifically, it first annotates the representative nodes and edges, which correspondently construct the annotation graph. Then GAGA facilitates the effective alignment of structures between the annotation graph and TAG. 

\textbf{LLaGA}~\cite{chen2024llagalargelanguagegraph}: To effectively translate graph structure into a language-based sequence for LLMs to process, LLaGA reorganizes the graph nodes to topology-aware sequence and map them into the token embeddings compatible with LLM input. LLaGA manages to extend to versatile datasets and tasks.

\textbf{Mol-MVMoE}~\cite{MolMvMOE}addresses the need for efficient and scalable models in chemical-based machine learning by leveraging a Multi-View Mixture of Experts (MoE) architecture. The approach enhances small molecule analysis by fusing embeddings from various chemical models and using a gating network to assign weights to different perspectives, improving robustness and accuracy.

\textbf{All in One}~\cite{AllInOne} addresses the challenge of applying pre-trained models to diverse graph tasks, where the pretext tasks often lead to negative transfer. The authors propose a novel multi-task prompting method for graph models, inspired by NLP's prompt learning, which unifies graph and language prompts and uses meta-learning to refine prompt initialization, improving model adaptability and performance across various graph tasks.

\textbf{GaLM}~\cite{xie2023graphawarelanguagemodelpretraining} addresses the gap in pre-training models on large heterogeneous graphs with textual information, as known as Large Graph Corpus, a step beyond previous studies focused on graphs alone. It combines large language models with graph neural networks and offers various fine-tuning methods for downstream applications. 

\textbf{OMOG}~\cite{OMOG} tackles the challenge of generalizing GNNs across different domains, where existing models require domain-specific architectures and training. OMOG pretrains a bank of expert models for specific datasets and uses gating functions to select and integrate expert knowledge, minimizing negative transfer when applied to new graphs.

\textbf{GraphGPT}~\cite{tang2024graphgptgraphinstructiontuning} is designed to enhance generalization in zero-shot learning environments for graph models, by LLMs with graph structural knowledge. The framework features a text-graph grounding component and a dual-stage instruction tuning approach, allowing LLMs to better understand and adapt to complex graph structures without relying on labeled downstream data.

\textbf{TouchUP-G}~\cite{zhu2023touchupgimprovingfeaturerepresentation} addresses the challenge of improving node features extracted from Pretrained Models for graph tasks, which are often graph-agnostic and limit GNN performance. TouchUp-G, a "Detect and Correct" approach, refines node features by detecting alignment issues using a novel feature homophily metric and correcting misalignments through a simple touchup on the Pretrained Models, enhancing GNN effectiveness.

\textbf{GraphTranslator}~\cite{zhang2024graphtranslatoraligninggraphmodel} bridges graph models and LLMs to handle both pre-defined and open-ended tasks. It integrates graph models with LLMs and a Producer to create graph-text alignment data, the approach allows LLMs to enhance node representations and make predictions based on language instructions, enabling a unified solution for diverse graph-related tasks.

\textbf{GraphText}~\cite{zhao2023graphtextgraphreasoningtext}: is a novel framework that translates graphs into natural language by deriving a graph-syntax tree encapsulating node attributes and inter-node relationships, enabling LLMs to treat graph tasks as text generation tasks. This approach allows for training-free graph reasoning, where GRAPHTEXT with ChatGPT can outperform supervised graph neural networks through in-context learning, and facilitates interactive graph reasoning, enabling seamless communication between humans and LLMs.

\textbf{PGT}~\cite{pgt}: Graph pre-training has primarily focused on small graphs or fixed graphs, but scaling it to industrial web-scale graphs with billions of nodes is challenging. To address this, the authors propose PGT (Pre-trained Graph Transformer), a scalable transformer-based framework with inductive abilities for making predictions on new nodes and graphs. Their framework, which uses a masked autoencoder architecture and a novel decoder feature augmentation strategy, has been successfully deployed on Tencent’s online game data, achieving state-of-the-art performance on both industrial and public datasets.

\textbf{SimTEG}~\cite{duan2023simtegfrustratinglysimpleapproach} is a simple approach for Textual Graph learning that combines supervised fine-tuning of a pre-trained language model with GNNs. By generating node embeddings from the fine-tuned Language Model’s hidden states, this method significantly enhances GNN performance on node classification and link prediction tasks across multiple graph benchmarks, without the need for complex frameworks or self-supervised tasks.

\textbf{InstructGraph}~\cite{wang2024instructgraphboostinglargelanguage} enhances LLMs for graph reasoning and generation tasks through instruction tuning and preference alignment. It introduces a structured verbalizer for unified graph representation, a tuning stage to guide graph tasks, and a preference alignment method to mitigate hallucination problems in graph tasks by optimizing model preferences using negative instance sampling.

\textbf{GOFA}~\cite{GOFA} is a generative graph language model built to create a Graph Foundation Model that combines self-supervised pretraining, task fluidity, and graph awareness. GOFA integrates GNN layers into a pre-trained LLM, enabling both semantic and structural modeling, and is pre-trained on graph-specific tasks, achieving strong performance in solving graph reasoning problems in zero-shot settings.

\textbf{MuseGraph}~\cite{tan2024musegraphgraphorientedinstructiontuning} addresses the limitation of traditional GNNs, which require re-training for different graph tasks. MuseGraph combines GNNs with LLMs to improve graph mining across diverse tasks and datasets. Key designs include adaptive input generation for compact graph descriptions, task-specific Chain-of-Thought-based instructions for reasoning, and dynamic instruction tuning to enhance model generalization and performance across tasks.

\textbf{SCORE}~\cite{SCORE} overcomes the limitations of current Graph Foundation Models, which often require extensive fine-tuning to handle new graphs and tasks. SCORE is a unified graph reasoning framework designed for zero-shot learning, to generalize across diverse graph tasks using Knowledge Graphs as a common topological structure. SCORE leverages semantic-conditional message passing to integrate both structural and semantic features, demonstrating significant performance improvements on 38 diverse datasets. 

\textbf{EdgePrompt}~\cite{anonymous2024edge} addresses the objective gap between self-supervised pre-training of GNNs and downstream tasks. EdgePrompt is a novel graph prompt tuning method that focuses on learning edge-specific prompts, rather than node features, to enhance the quality of graph representations. By incorporating these edge prompts through message passing in pre-trained GNNs, EdgePrompt effectively captures structural information and is compatible with various GNN architectures, improving performance across different tasks.

\textbf{GFSE}~\cite{anonymous2024gfse} addresses the limitations of existing pre-trained graph models, which are often domain-specific and struggle to capture complex graph structures. GFSE is a cross-domain model designed to identify universal structural patterns across diverse graph domains, such as molecular structures and social networks. GFSE, built on a Graph Transformer with attention mechanisms, produces expressive positional and structural embeddings that can be integrated with various downstream models, significantly improving performance with minimal fine-tuning.

\textbf{GIT}~\cite{GIT} addresses the challenge of discovering shared patterns (generalities) across graphs, which has hindered the development of effective graph foundation models. The authors propose a novel approach using task trees to capture shared generalities in graphs, theorizing that these patterns are preserved in the task trees of graphs. Their proposed model is pre-trained on diverse task trees and demonstrates strong performance in fine-tuning, in-context learning, and zero-shot learning across multiple domains, showing its ability to generalize and outperform domain-specific models.

\textbf{GL-Fusion}~\cite{anonymous2024glfusion} addresses the limitations of integrating LLMs and GNNs, where LLM-centered models struggle with graph structure and GNN-centered models lose semantic information from textual data. GL-Fusion integrates GNNs with LLMs through three key innovations: Structure-Aware Transformers for joint processing of textual and structural information, Graph-Text Cross-Attention for uncompressed semantic integration, and a GNN-LLM Twin Predictor to enable flexible autoregressive generation and scalable prediction.

\textbf{OFA}~\cite{OFA}: This research is to create a unified model that can handle diverse graph tasks, overcoming the challenges of distinct graph data attributes, task diversity, and the lack of an effective graph prompting paradigm. OFA uses text-attributed graphs to unify different graph data types and applies language models to encode these attributes into a shared embedding space. OFA introduces "nodes-of-interest" to standardize tasks and a novel graph prompting paradigm that allows for in-context learning without fine-tuning, demonstrating strong performance across multiple domains in supervised, few-shot, and zero-shot learning scenarios.

\textbf{WalkFM}~\cite{WalkLM} aims to address the challenges of modeling complex attributes and structures in real-world graphs, in which traditional GNNs often require extensive training for specific tasks. WalkLM integrates language models with random walks to learn unsupervised, generic graph representations, capturing both attribute semantics and graph structures. By performing attributed random walks and fine-tuning a Language Model on the generated textual sequences, their approach achieves significant improvements in node embeddings for various downstream tasks, offering a more data-efficient and flexible solution than existing unsupervised methods.

\textbf{RAGraph}~\cite{jiang2024ragraphgeneralretrievalaugmentedgraph} enhance the generalization capabilities of GNNs with unseen graph data. It incorporates external graph data through a toy graph vector library, which captures key attributes for improving model generalization. During inference, It retrieves similar toy graphs and integrates them via a message-passing prompting mechanism, achieving superior performance in various downstream tasks without requiring task-specific fine-tuning.

\textbf{GraphPrompt}~\cite{liu2023graphpromptunifyingpretrainingdownstream} reduces the reliance on task-specific supervision in GNNs by utilizing pre-training and prompting techniques, which have been effective in natural language processing. GraphPrompt unifies pre-training and downstream tasks into a common task template and uses learnable prompts to help the model retrieve the most relevant knowledge for each specific task. This approach improves GNN performance across various downstream tasks while minimizing the need for extensive task-specific labeling.

\textbf{MultiGPrompt}~\cite{yu2024multigpromptmultitaskpretrainingprompting} improves the performance of GNNs in few-shot settings and reduces reliance on task-specific labels by leveraging multi-task pre-training and prompting. MultiGPrompt uses multiple pretext tasks during pre-training to capture more comprehensive knowledge and introduces a dual-prompt mechanism—composed and open prompts—to guide downstream tasks. 

\textbf{AnyGraph}~\cite{xia2024anygraphgraphfoundationmodel} is a unified graph model designed to address key challenges in graph learning: structure and feature heterogeneity, fast adaptation to new domains, and scaling law emergence. Built on a Graph Mixture-of-Experts (MoE) architecture, AnyGraph efficiently handles distribution shifts, enables quick adaptation through a lightweight expert routing mechanism, and demonstrates favorable scaling with data and model size.

\textbf{GPT4Graph}~\cite{guo2023gpt4graphlargelanguagemodels} presents an empirical study evaluating the ability of LLMs to understand and reason about graphs. By assessing their performance across a variety of structural and semantic tasks, this work identifies current limitations and suggests future directions for improving LLMs' proficiency in handling graph-related challenges.

\textbf{GraphAlign}~\cite{hou2024graphalignpretraininggraphneural} is designed to address the challenge of feature discrepancy in graph self-supervised learning. By aligning feature distributions across diverse graphs using encoding, normalization, and MoE modules, GraphAlign enables effective pre-training of a unified GNN that performs well on cross-domain graphs.

\textbf{OpenGraph}~\cite{xia2024opengraphopengraphfoundation} is a novel graph foundation model designed to address the challenge of generalizing to unseen graph data with different properties. By enhancing data augmentation with an LLM, employing a unified graph tokenizer for better generalization, and using a scalable graph transformer to capture node-wise dependencies, OpenGraph achieves impressive zero-shot graph learning performance across diverse settings.

\textbf{Prodigy}~\cite{huang2023prodigy} is the first pre-training framework enabling in-context learning over graphs, where pre-trained models adapt to new tasks without parameter optimization. By using a novel prompt graph representation and a graph neural network, it achieves strong performance on tasks, like citation networks and knowledge graphs, outperforming contrastive pre-training baselines by 18 percent and standard fine-tuning with limited data by 33 percent on average.

\textbf{GraphAgent-Reasoner}~\cite{hu2024scalableaccurategraphreasoning} overcomes the limitations of LLMs in handling complex graph reasoning tasks, particularly when dealing with intricate graph structures and long texts. GraphAgent-Reasoner is a fine-tuning-free framework that employs a multi-agent collaboration strategy to decompose graph problems into smaller, node-centric tasks, which are distributed among agents to improve reasoning accuracy. The framework efficiently handles larger graphs and achieves near-perfect accuracy on polynomial-time graph reasoning tasks.

\textbf{Talk Like a Graph}~\cite{fatemi2023talklikegraphencoding} aims to address the challenges of reasoning on graphs using LLMs, an area that has been underexplored despite advancements in automated reasoning with natural text. The authors conduct the first comprehensive study on encoding graph-structured data as a text for LLMs, revealing that LLM performance on graph reasoning tasks is influenced by three key factors: the graph encoding method, the nature of the task, and the structure of the graph itself. These insights offer valuable guidance to improve LLM-based graph reasoning.

\textbf{NLGraph}~\cite{NLGraph} explores whether LLMs can explicitly process textual descriptions of graphs, map them to grounded conceptual spaces, and perform structured operations. NLGraph is a comprehensive benchmark with 29,370 problems across eight graph reasoning tasks of varying complexity, evaluating LLMs' graph reasoning capabilities. Authors find that while LLMs show some ability in graph reasoning, their performance deteriorates on more complex tasks and when faced with spurious correlations, leading to the proposal of two new prompting techniques—Build-a-Graph and Algorithmic Prompting—to improve LLMs in solving natural language graph problems.

\textbf{GraphLLM}~\cite{chai2023graphllmboostinggraphreasoning}: Despite the remarkable progress of  LLMs towards AGI, they still struggle with graph reasoning tasks due to the common reliance on converting graphs into natural language descriptions. To address this, this work proposes GraphLLM, an end-to-end approach that integrates graph learning models with LLMs, enabling LLMs to effectively interpret and reason on graph data, with strong empirical results across four key graph reasoning tasks.

\textbf{GraphToken}~\cite{perozzi2024letgraphtalkingencoding} is a parameter-efficient method for encoding structured data into a sequential form suitable for LLMs. By explicitly representing graph structure in prompts, GraphToken significantly improves performance across various reasoning tasks. 

\textbf{LINK2Doc}~\cite{Link2Doc} is a novel framework for link prediction on textual-edge graphs. By summarizing neighborhood information between node pairs as human-written documents, LINK2DOC preserves both semantic and topological details, enhancing the graph neural network's text-understanding ability through self-supervised learning, and outperforming existing edge-aware GNNs and language models in link prediction tasks across four real-world datasets.

\textbf{UniGraph}~\cite{he2024unigraphlearningunifiedcrossdomain} is a foundation model framework designed for TAGs, which leverages textual features to enable generalization across diverse graph domains and tasks. By combining Language Models and GNNs in a novel cascaded architecture, UniGraph achieves zero-shot and few-shot transfer capabilities, outperforming traditional GNNs trained on specific datasets, and demonstrating effective self-supervised learning on unseen graphs through a Masked Graph Modeling pre-training algorithm.

\section{Classifying Based on Downstream Tasks: The Role of Applications}

In the research of Graph Neural Networks (GNNs) and Large Language Models (LLMs), the ultimate goal is to enable these models to be effectively applied to real-world tasks. Thus, classifying existing studies from the perspective of downstream tasks holds significant importance. Such a classification not only clarifies the application scope of the models but also guides the design of pretraining mechanisms, especially in scenarios with unlabeled data, by constructing unsupervised training frameworks that serve as a foundation for downstream applications.

In practical applications, models need the capability to effectively model generic graph-centric knowledge, which is often achieved through unsupervised pre-training. Therefore, we classify existing research along two key dimensions: training mechanisms and application scenarios. The training mechanism dimension focuses on whether the learning is supervised or unsupervised, while the application scenario dimension examines whether the model is suited for single-domain or multi-domain applications.

For instance, in real-world applications, if a model has access to a small amount of labeled data, it can be efficiently fine-tuned to achieve rapid adaptation. Conversely, in the absence of labeled data, the model relies on efficient inference mechanisms to fully exploit its generalization capability. In terms of application scenarios, it is essential to determine whether the model is intended for a single-task setting (e.g., citation networks) or designed to generalize across multiple tasks (e.g., bioinformatics, and social networks). Moreover, the mechanism for updating model parameters is a critical consideration, including whether updates are based on supervised learning with labeled data or unsupervised self-adaptive approaches.

Based on these classification criteria, we systematically categorize existing research into the following five categories, which depict how each research trains and selects targeted scenarios. The comprehensive summaries are as follows:

\begin{enumerate}
    \item \textbf{Single-Domain \& Supervised Learning (Fine-tuning):} This category focuses on specific tasks within a single domain, utilizing a small amount of labeled data for fine-tuning. The representative methods are as follows: LPNL\cite{bi2024lpnlscalablelinkprediction}, DGTL\cite{qin2024disentangledrepresentationlearninglarge}, GraphLLM\cite{chai2023graphllmboostinggraphreasoning}, SimTeG\cite{duan2023simtegfrustratinglysimpleapproach}, AUGGLM\cite{xu2024languagemodelsgraphlearners}, GQT\cite{wang2024learninggraphquantizedtokenizers}, HIGHT\cite{chen2024highthierarchicalgraphtokenization}, RoSE\cite{seo2024unleashingpotentialtextattributedgraphs}, CurGL\cite{anonymous2024curriculum}, DP-GPL\cite{anonymous2024dpgpl}, InstructGLM\cite{Instruct-GLM}, GRAD\cite{mavromatis2023traingnnteachergraphaware}, GraphBridge\cite{anonymous2024graphbridge}, GAGA\cite{GAGA}, GraphProp\cite{anonymous2024graphprop}, GLEM\cite{zhao2023learninglargescaletextattributedgraphs}, GIANT\cite{chien2022nodefeatureextractionselfsupervised}, TAPE\cite{he2024harnessingexplanationsllmtolminterpreter}, LLaGA\cite{chen2024llagalargelanguagegraph}, Mol-MVMoE\cite{MolMvMOE}, ENGINE\cite{zhu2024efficienttuninginferencelarge}, HiGPT\cite{tang2024higptheterogeneousgraphlanguage}, LMGJoint\cite{SheaFormer}, GPF\cite{fang2023universal}, PromptGFM\cite{PromptGFM}, SheaFormer\cite{SheaFormer}, GraphGPT\cite{tang2024graphgptgraphinstructiontuning}, TouchUp-G\cite{zhu2023touchupgimprovingfeaturerepresentation}, SKETCH\cite{huang2024gnngoodadapterllms}, GraphAdapter\cite{huang2024gnngoodadapterllms}, GraphTranslator\cite{zhang2024graphtranslatoraligninggraphmodel}, \\ GraphPrompter\cite{graphprompter}, GPPT\cite{gppt}.

    \item \textbf{Single-Domain \& Unsupervised Learning:} This category targets single-domain settings and employs unsupervised learning mechanisms to extract generic graph features from unlabeled data. The research work mainly includes: PATTON\cite{jin2023pattonlanguagemodelpretraining}, ConGraT\cite{brannon2024congratselfsupervisedcontrastivepretraining}, FineMolTex\cite{li2024finemoltexfinegrainedmoleculargraphtext}, Link2Doc\cite{Link2Doc}, PGT\cite{pgt}, TAGExplainer\cite{pan2024tagexplainernarratinggraphexplanations}, TAGA\cite{pan2024distillinglargelanguagemodels}, Grenade\cite{li2023grenadegraphcentriclanguagemodel}, THLM\cite{THLM}, ZeroG\cite{li2024zeroginvestigatingcrossdatasetzeroshot}, CIKM-KD\cite{pan2024distillinglargelanguagemodels}, GAugLLM\cite{fang2024gaugllmimprovinggraphcontrastive}, GaLM\cite{xie2023graphawarelanguagemodelpretraining}, OMOG\cite{OMOG}.

    \item \textbf{Multi-Domain \& Supervised Learning (Fine-tuning):} This category explores the application of supervised fine-tuning across multiple data domains, enabling the model to adapt and extend to multi-task practical application environments. Currently, the main works in this category are shown as follows: InstructGraph\cite{wang2024instructgraphboostinglargelanguage}, GraphBridge (ICLR)\cite{anonymous2024graphbridge}, All in One\cite{AllInOne}, EdgePromp\cite{anonymous2024edge}, GraphPrompt\cite{liu2023graphpromptunifyingpretrainingdownstream}, MultiGPrompt\cite{yu2024multigpromptmultitaskpretrainingprompting}, MuseGraph\cite{tan2024musegraphgraphorientedinstructiontuning}, RAGraph\cite{jiang2024ragraphgeneralretrievalaugmentedgraph}, GraphFM (ICLR)\cite{anonymous2024graphfm},  GIT\cite{GIT}.

    \item \textbf{Multi-Domain \& Unsupervised Learning:} This category emphasizes the generalization capability of unsupervised learning mechanisms, ensuring robust performance across diverse domains and tasks. These papers are the following: AnyGraph\cite{xia2024anygraphgraphfoundationmodel}, GOFA\cite{xia2024anygraphgraphfoundationmodel}, GraphAny\cite{zhao2024graphanyfoundationmodelnode}, SCORE\cite{xia2024anygraphgraphfoundationmodel}, UniGraph\cite{he2024unigraphlearningunifiedcrossdomain}, OpenGraph\cite{xia2024opengraphopengraphfoundation}, GFSE\cite{anonymous2024gfse}, GL-Fusion\cite{anonymous2024glfusion}, OFA\cite{OFA}, PRODIGY\cite{huang2023prodigy}, WalkLM\cite{WalkLM}, GraphAlign\cite{hou2024graphalignpretraininggraphneural}.

    \item \textbf{Few-Shot and Zero-Shot Inference:} This category investigates efficient inference techniques in scenarios with limited or no labeled data, fully leveraging the generalization and flexibility of pre-trained models. The main works in this category include: GPT4Graph\cite{guo2023gpt4graphlargelanguagemodels}, GraphText\cite{zhao2023graphtextgraphreasoningtext}, GraphAgent-Reasoner\cite{hu2024scalableaccurategraphreasoning}, GraphToken\cite{perozzi2024letgraphtalkingencoding}, Talk like a Graph\cite{fatemi2023talklikegraphencoding}, NLGraph\cite{NLGraph}, LLM-GNN\cite{chen2024labelfreenodeclassificationgraphs}.

\end{enumerate}

Through the above classification framework, we not only systematically analyze existing research from the perspectives of training mechanisms and application scenarios but also uncover its relevance and contributions to real-world tasks. The researcher should select pre-training methodologies according to desirable downstream tasks. For acquiring a more generalizable knowledge across multi-domain applications, the concept of unsupervised learning and alignment architectures of GNNs and LLMs training should be emphasized. 

\section{Training Paradigm for applying LLM-GNN to realistic challenges}

Integrating GNNs and LLMs offers a powerful three-step paradigm—pre-training, fine-tuning, and inference—for tackling complex real-world problems. In the pre-training phase, GNNs learn generalizable representations by leveraging techniques such as message-passing or graph embeddings to capture structural and relational patterns from graph data. In parallel, LLMs utilize transformer architectures and attention mechanisms to extract semantic understanding from unstructured text. This step is often unsupervised or self-supervised, leveraging massive datasets to acquire the generalizable knowledge to be transferred to the next two training phases.  Next, during fine-tuning, the pre-trained models are adapted to specific tasks, such as link prediction, node classification, or knowledge-enhanced reasoning, using task-specific data. Fine-tuning often involves integrating GNN and LLM outputs through mechanisms like cross-modal attention layers or feature fusion, aligning the strengths of GNNs in relational reasoning with the contextual capabilities of LLMs. In the inference phase, the integrated model applies its learned capabilities collaboratively to make predictions or generate outputs on new, unseen data. Depending on the availability of data and the nature of tasks, training methodologies such as supervised learning, few-shot, and zero-shot inference can be applied. 

\subsection{Pre-training phase}

Pre-training in the integration of LLMs and GNNs involves leveraging diverse training methodologies—unsupervised learning, contrastive learning, and generative self-supervised learning—to acquire the generalizable graph knowledge before training the models with datasets with any labeled information. The key idea is to leverage GNNs' ability to model and learn structural information, which can be transferred to downstream tasks during the fine-tuning phase. Similarly, pre-training LLMs enhance their capacity for understanding text, which serves as a foundation for fine-tuning.

Unsupervised learning in GNNs often focuses on graph-based tasks like node embedding generation using random walk-based methods (e.g., DeepWalk, Node2Vec) or message-passing frameworks, which capture structural and relational information. In parallel, LLMs utilize massive unstructured text corpora to learn semantic patterns through autoregressive objectives (e.g., GPT models) or masked language modeling (e.g., BERT). Contrastive learning bridges both modalities by aligning representations from graph and text data. For instance, nodes in a graph and their textual descriptions can be aligned via contrastive losses, enhancing cross-modal understanding for tasks like knowledge graph completion or entity alignment. Generative self-supervised learning further deepens the integration, where GNNs predict masked nodes or edges while LLMs predict masked tokens, enabling shared pre-training over hybrid graph-text datasets. 
This diverse suite of pre-training methods ensures that the integrated models capture both relational complexity and rich contextual semantics, setting a strong foundation for downstream tasks.

\subsection{Fine-tuning Phase}

The fine-tuning phase tailors pre-trained models to specific tasks through specific guidance referred as task-specific prompts and instructions. Specifically, the pre-trained encoder remains frozen as it has already learned useful general features. Instead, only the parameters of the projectors (such as classifiers or projection heads) are trainable, and they are fine-tuned to adapt the model's representations to the specific downstream task. Prompt-based fine-tuning involves designing specific input prompts that guide the model's behavior toward solving a particular task. For example, in the context of graph-related tasks, a prompt such as “Predict the missing links between these nodes in the graph” can direct the GNN model to focus on link prediction, while a prompt like “Given a graph and text, answer the query based on the combined information” can guide both GNNs and LLMs to handle tasks that require integrated reasoning from both graph structure and textual content.

On the other hand, instruction-based fine-tuning provides more explicit guidance by framing the task with clear instructions on how the model should behave. For example, an instruction might be “Classify the nodes in this graph based on their roles (e.g., user, admin)” for GNNs, or “Given this paragraph, summarize the key points” for LLMs.

Prompt-based and instruction-based fine-tuning complement each other by balancing flexibility and structure in multi-modal tasks. Prompt-based fine-tuning offers adaptability, guiding the model to focus on specific tasks, while instruction-based fine-tuning ensures a clear framework for integrating diverse data sources. Together, they improve task alignment by helping the model understand how to combine structured knowledge (from graphs) and unstructured knowledge (from text), which is essential for tasks like knowledge-enhanced reasoning.

\subsection{Inference Phase}

The inference phase is the final stage where the fine-tuned model is applied to make predictions or generate outcomes for new, unseen data. This phase can leverage different methodologies depending on the availability of labeled data and the nature of the task. The primary inference strategies—supervised learning, few-shot inference, and zero-shot inference—are adapted to the integrated model to handle various multi-modal tasks involving both graph and text data.

In the context of supervised learning, the integrated model applies its learned knowledge from both the GNN and the LLM to make predictions based on labeled data. After fine-tuning, the model has learned task-specific patterns and can now generate predictions for new inputs. For example, in a task that requires both graph-based relational reasoning and textual understanding, the model might predict the relationship between entities in a graph and then use the fine-tuned LLM to generate a textual explanation of that relationship. Similarly, for tasks such as node classification or link prediction, the model can combine the graph embeddings from the GNN with the textual context from the LLM to make task-related predictions.

For few-shot inference, the integrated model can perform, such as similarity matching of embeddings derived from both the graph and textual components, even with very few labeled examples. Here, the GNN extracts graph embeddings by understanding node relationships and subgraph structures, while the LLM generates embeddings that capture the semantic content of textual data. In the inference phase, when given a query or a new example with minimal labeled data, the model compares the embeddings of the new input to those learned during fine-tuning, determining the closest match. For instance, in a multimodal recommendation task, the model might match a user’s query to a set of products by comparing both the user-item interaction graph and the textual product descriptions. This allows the model to generate recommendations or answers based on the similarity of embeddings, despite only having a few labeled examples.

In zero-shot inference, the integrated model can match labels or generate predictions based on its semantic understanding of both graph structures and text, without needing specific labeled examples for the task at hand. The model leverages its pre-trained and fine-tuned knowledge to infer answers or labels by understanding the broader meaning behind the data. For example, in a graph-based question-answering task where the model must answer a query by combining graph structure and textual content, the model could predict the answer by reasoning about the graph's relational patterns and using the LLM's semantic understanding to interpret and match the query to relevant knowledge, all without having seen specific examples during fine-tuning.

\section{Conclusion}

This study presents a novel classification framework based on the three core elements of machine learning—data, model, and downstream task—to systematically categorize and analyze the integration of Graph Neural Networks (GNNs) and Large Language Models (LLMs). Through this structured review, we summarize recent advancements, challenges, and potential research directions in this emerging field. The key contributions and findings of this work are as follows:
\begin{enumerate}
\item \textbf{Proposal of a Unified Classification Framework.}
We introduce a new framework that organizes existing research by the three pillars of machine learning: data, model, and downstream task. This framework serves as a theoretical foundation for understanding and synthesizing efforts in integrating graph data and language models, offering a comprehensive lens to explore their interplay across diverse domains and applications.

\item \textbf{Comprehensive Analysis of Cross-Domain Tasks.}
By categorizing research into five perspectives—Single-task \& Single-domain, Single-task \& Multi-domain, Multi-task \& Single-domain, Multi-task \& Multi-domain, and Graph Reasoning, we provide an in-depth analysis of the diversity and challenges. Our findings reveal that as the complexity of tasks and domains increases, the synergy between GNNs and LLMs becomes critical, laying the groundwork for enhancing model generalizability in multi-task and cross-domain settings.

\item \textbf{Exploration of Model Collaboration Strategies.}
We identify five collaborative paradigms for integrating GNNs and LLMs: independent modules, GNN-enhanced LLMs, LLM-enhanced GNNs, standalone GNNs, and standalone LLMs. We analyze how these paradigms address the challenges of learning from graph data enriched with textual attributes. Our findings indicate that in tasks that require the simultaneous handling of complex graph structures and rich textual information, GNNs and LLMs can complement each other, jointly enhancing the performance of graph learning tasks. This synergy effectively overcomes the limitations of traditional GNN models in cross-domain scenarios, enabling more flexible and efficient graph learning, and ultimately advancing the realization of a Graph Foundation Model.

\item \textbf{Exploration of Cross-Domain Training Mechanisms.}
We categorize the research based on the differences in training mechanisms in real-world application scenarios: Single-Domain and Supervised Learning, Single-Domain and Unsupervised Learning, Multi-Domain and Supervised Learning, Multi-Domain and Unsupervised Learning, and Few-Shot and Zero-Shot Inference. We analyze the advantages and limitations of each category of methods, highlighting the relevance of training mechanisms to application scenarios. This classification framework provides a structured understanding of how various training strategies can address specific challenges in graph-centric tasks, inspiring further advancements in the Graph Foundation Model.

\end{enumerate}

\newpage
\bibliographystyle{plain} 
\bibliography{reference} 

\end{document}